\documentclass[preprint,12pt]{elsarticle}

\usepackage{amsmath}
\usepackage{amssymb}
\DeclareMathOperator*{\argmax}{arg\,max}
\DeclareMathOperator*{\argmin}{arg\,min}

\usepackage{algorithm}
\usepackage{algpseudocode}
\algrenewcommand\algorithmicrequire{\textbf{Input:}}
\algrenewcommand\algorithmicensure{\textbf{Output:}}

\journal{Journal of The Franklin Institute}

\begin{document}

\begin{frontmatter}



\title{Probabilistic Constraint Tightening Techniques for Trajectory Planning with Predictive Control}


\author[CU]{Nathan Goulet\corref{cor1}}
\cortext[cor1]{Corresponding Author}
\ead{ngoulet@clemson.edu}
\author[Huawei]{Qian Wang}
\author[CU]{Beshah Ayalew}
\ead{beshah@clemson.edu}

\address[CU]{Department of Automotive Engineering, Clemson University, 4 Research Dr., Greenville, 29607, SC, USA}
            
\address[Huawei]{Autonomous Driving Solutions (ADS), Huawei, Shanghai, China}

%

\begin{abstract}
In order for automated mobile vehicles to navigate in the real world with minimal collision risks, it is necessary for their planning algorithms to consider uncertainties from measurements and environmental disturbances. In this paper, we consider analytical solutions for a conservative approximation of the mutual probability of collision between two robotic vehicles in the presence of such uncertainties. Therein, we present two methods, which we call unitary scaling and principal axes rotation, for decoupling the bivariate integral required for efficient approximation of the probability of collision between two vehicles including orientation effects. We compare the conservatism of these methods analytically and numerically. By closing a control loop through a model predictive guidance scheme, we observe through Monte-Carlo simulations that directly implementing collision avoidance constraints from the conservative approximations remains infeasible for real-time planning. We then propose and implement a convexification approach based on the tightened collision constraints that significantly improves the computational efficiency and robustness of the predictive guidance scheme.
\end{abstract}




\end{frontmatter}


\section{Introduction}\label{sec:Intro}

A core undertaking for safely guiding an autonomously controlled vehicle (ACV) within the real world amongst other dynamic agents, objects or object vehicles (OVs), is that of devising real-time feasible collision free motion plans. Guidance and trajectory planning algorithms for an ACV operating in dynamic environments need to observe/track and predict the motion and extent of neighboring OVs \cite{Lefevre2014}. Starting with available measurements such as sparse laser point clouds \cite{Wender2008}, OVs can be tracked by treating each as an extended object \cite{Mihaylova2014,Granstrom2016}. With knowledge of an OV’s shape and motion states (for example fusing with camera images and other sensors),  it can be represented by an estimated geometric/spatial shape to formulate enforceable collision avoidance constraints  in the motion trajectory planning problem for the ACV \cite{Goerzen2010,Paden2016}. However, the problem is still quite challenging to solve in real-time due to the following principal difficulties: 1) The planning problem involves uncertainties due to modeling errors, sensor imperfections and/or environmental disturbances; and; 2) Since the desired feasible field for planning maneuvers is defined to be outside of the area occupied by OVs and other (road or workspace) boundaries, the collision avoidance problem is generally non-convex.

The first difficulty involving uncertainties is often handled by considering the known/assumed bounds (non-deterministic uncertainty case) or their probability distributions (probabilistic case) \cite{Mesbah2016}. In the non-deterministic case \cite{Kuwata2005}, the worst case of the uncertainty is considered in the trajectory planning problem leading often to a very conservative solution. In the probabilistic case \cite{Blackmore2006}, the conservatism can be relaxed by specifying probabilistic collision criteria which are often posed as chance/probabilistic collision avoidance constraints with a given threshold. However, the evaluations of these collision criteria are often computationally intractable due to the multivariate integrals involved in the probabilistic collision avoidance criteria. Nevertheless, for a specified confidence threshold, a solution can be sought by solving approximate deterministic trajectory planning problems with either a sampling-based method (Monte Carlo Simulations) \cite{Blackmore2006} or with constraint-tightening methods \cite{Wang2017a,Carvalho2014}.

In constraint-tightening methods, the original probabilistic collision avoidance (chance) constraint is replaced by a deterministic constraint that is a function of the confidence threshold and the probability distribution of the uncertainty. The key step in these methods is to determine the form of this function. With assumptions of uncorrelated Gaussian distributions for the state observations, an explicit function is easy to obtain by factorizations into univariate integrals \cite{Carvalho2014}. For the case of correlated state uncertainties, \cite{DuToit2011} derived an approximate explicit function for probabilistic collision evaluation of small-sized objects (radius smaller than 1 m). However, this approximation does not work for agents of larger geometric sizes, e.g. ACVs operating in public road traffic. Our prior work in \cite{Wang2017a} provided a framework to handle agents of non-negligible sizes with 2D rectangular axes-aligned shapes. Therein, the computations of the mutual probability of collision between OVs of non-negligible sizes is approximated analytically via coordinate rotations. In this paper, we will refer to this method as principal axes rotation. A drawback of this framework is that it does not account for the relative orientation of agents and the associated uncertainty. This limitation will be addressed in this paper by drawing from a similar approach proposed in \cite{Hardy2013} that accounts for heading state uncertainty assuming independence of the heading state from the translational states. Therein, the mutual probability of collision is approximated via a linear coordinate transformation to achieve unit covariance. In this paper, we will refer to this latter method as unitary scaling. Both principal axes rotation and unitary scaling methods attempt to achieve the same goal of constraint tightening via conservative approximations of the collision avoidance criteria. In this paper, we characterize the conservatism introduced by each method to identify guidance on their respective preferred use scenarios.

Approaches to address the second difficulty (non-convexity of planning space) vary by the choice of the planning method. Commonly cited sampling-based planning methods like the A* algorithm \cite{Kushleyev2009} and RRT* algorithm \cite{Kuwata2009} use polygonal models \cite{Blackmore2006,Nilsson2015}, which are a disjunction of linear constraints, or algebraic models like circles, ellipses \cite{Rosolia2015} and hyper ellipses \cite{Menon2015}. These sampling-based methods suffer from the computational burden of the large number of samples that may be needed for sufficient resolution of the relevant geometry for real-time guidance in complex traffic. Alternatives to these planning methods are mathematical constrained-optimization based planning methods like model predictive control (MPC) \cite{Weiskircher2017} that solve on-line guidance optimizations subject to the constraints of the measured and model predicted evolution of traffic. While polygonal representations of objects can be adopted with these methods, they are less convenient since the disjunction of linear constraints that would be needed to represent multi-vehicle traffic often lead to non-smoothness and/or discontinuity in the solution space. This often result in Disjunctive Linear Programming problems \cite{Balas1979}, which require specific solvers to find a solution. On the other hand, if one finds analytical approximations for the mutual probability of collision criteria between objects, it is possible to directly implement those as tightened constraint functions in the motion trajectory planning task (as in \cite{Hardy2013}). However, this may still result in excessive execution time and non-robust behavior, as we will show later in Section \ref{subsec:direct} with an MPC-based planner. For this reason, we further propose convexification steps on top of the analytical approximations that improve execution time while still guaranteeing the probability of collision threshold is not violated by the planner.

The main contributions of this paper can be summarized as follows: 
\begin{itemize}
\item Detailed discussions on the analytical approximation of the mutual probability of collision avoidance criteria between two agents.
\item Compare two methods for decoupling the bivariate integral that appears in the collision avoidance criteria.
\item Demonstration of the shortcomings of a chance-constrained nonlinear MPC-based motion trajectory planner that directly implements the approximated probabilistic collision avoidance criteria as tightened constraints.
\item Propose a constraint convexification method that significantly enhances the computational speed and robustness of predictive planners that use the constraint-tightening approximations.
\end{itemize}
The rest of the paper is organized as follows. Section \ref{sec:criteria_approximation} derives the two methods for analytical approximation of the probability of collision criteria and characterizes the associated conservatism in each. Section \ref{sec:closed_loop} first outlines the chance-constrained nonlinear MPC-based motion-trajectory planner and demonstrates its performance with a direct implementation of the results from Section \ref{sec:criteria_approximation}. Then, it proposes the convexification method and demonstrates the associated improvements. Section \ref{sec:conclusion} presents the conclusions of the paper and identifies some avenues for further work.

\section{Conservative Approximation of Probabilistic Collision Avoidance Criteria}\label{sec:criteria_approximation}

\begin{figure}[!t]
\centering
\includegraphics[width=3.5in]{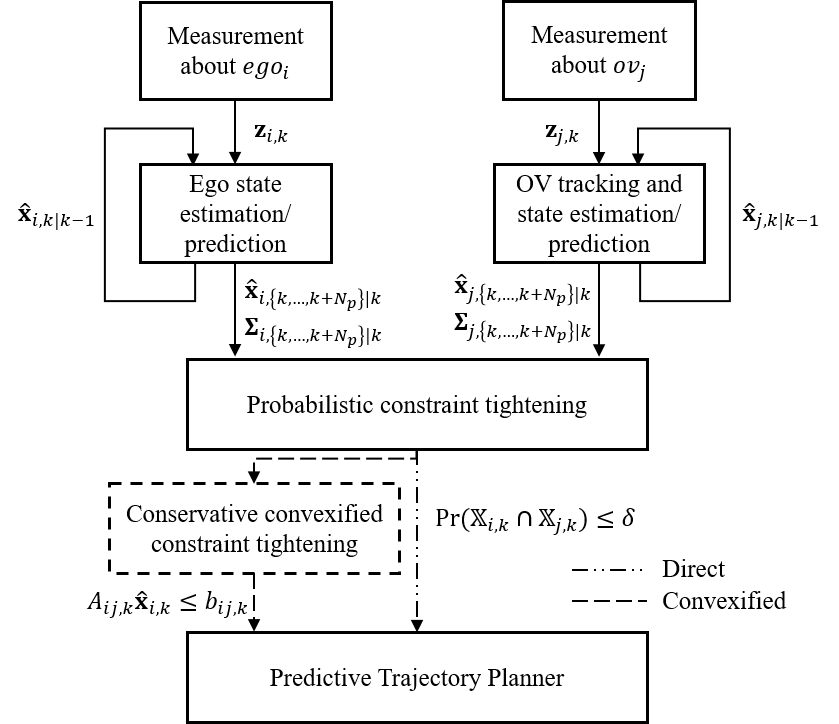}
\caption{Probabilistic collision avoidance constraint tightening framework.}
\label{fig:framework}
\end{figure}
Figure \ref{fig:framework} shows a schematic of the overall process for evaluating probabilistic collision avoidance constraints. We assume that there are measurements $\mathbf{z}_j$ available for each object vehicle, ${ov}_j$, in the field of view of the ego vehicle, ${ego}_i$. We also assume that each ${ov}_j$ is tracked from these measurements and we can obtain an estimation of the current state and prediction of the future states of each ${ov}_j$ and ${ego}_i$. As object vehicle tracking, and state estimation and prediction are not a primary contribution of this paper we omit further discussion of that here. The reader is referred to \cite{Wang2018,Petrovskaya2009,Yoon2019} and references therein. We note that estimated/predicted state vector $\hat{\mathbf{x}}_j$ and associated covariance matrices $\Sigma_j$ are assumed to be available for each ${ov}_j$ over the time horizon $N_p$. Using this information about the state and uncertainty of ${ov}_j$, along with the estimated and predicted state trajectory and covariance of ${ego}_i$, the probabilistic constraint tightening block in Figure \ref{fig:framework} conducts the main computations we detail in the next subsection. In Section \ref{sec:closed_loop}, we will first overview how to directly implement the probabilistic constraint  $\Pr\left(\mathbb{X}_{i,k}\cap\mathbb{X}_{j,k}\right)\leq\delta$ within a predictive trajectory planner (dash-dotted line in Figure \ref{fig:framework}), where $\mathbb{X}$ is the 2D space occupied by a vehicle considering its geometric shape and $\delta$ is the probability of collision threshold. Then, we will present our conservative convexification method (dashed lines in Figure \ref{fig:framework}). Either the direct probabilistic constraint or the convexified constraint ($A_{ij,k}\hat{x}_{i,k}\leq b_{ij,k}$) may be applied over the predicted state trajectories with the time indexed by $k$ as in Figure \ref{fig:framework}.

We will begin by formulating a conservative approximation (upper bound) on the probability of collision between ${ego}_i$ and ${ov}_j$. For our purposes, hereafter, $x$ represents the 2D position state for the centroid of the geometric shape for a given vehicle, unless specifically noted otherwise.

\subsection{General Formulation of the Probabilistic Collision Avoidance Criteria}\label{subsec:general}
\begin{figure}[!t]
\centering
\includegraphics[width=3.25in]{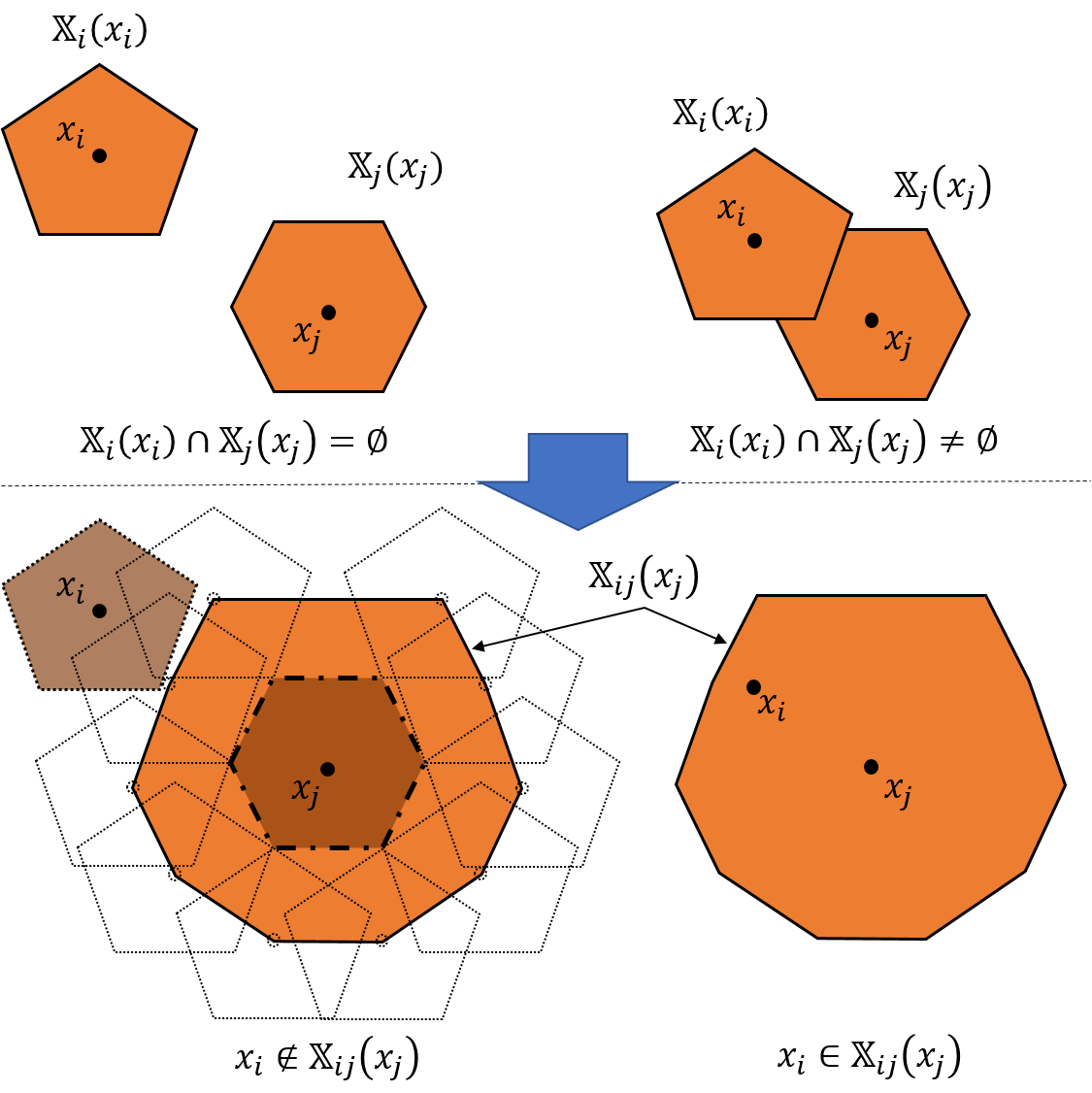}
\caption{Example of a collision free condition (left) and collision condition (right) between an ego vehicle (pentagonal here) ${ego}_i$ and object vehicle (hexagonal) ${ov}_j$.}
\label{fig:collision_def}
\end{figure}
We can state that a collision occurs between ${ego}_i$ and ${ov}_j$, if $\mathbb{X}_i\left(x_i\right)\cap\mathbb{X}_j\left(x_j\right)\neq\emptyset$ (at any time $k$, where the subscript $k$ is omitted in the following to reduce clutter). This condition is depicted on the right-hand side of Figure \ref{fig:collision_def}.  Any arbitrary shape may be considered for the vehicles at this point. The criteria for the probability of ${ego}_i$ entering a collision state with ${ov}_j$ to not exceed a threshold $\delta$  is then:
\begin{equation}\label{eq:PrC_intersect}
\Pr\left(\mathbb{X}\left(x_i\right)\cap\mathbb{X}\left(x_j\right)\neq\emptyset\right)\leq\delta.
\end{equation}
Determining the intersection of two arbitrary shapes, $\mathbb{X}_i\left(x_i\right)\cap\mathbb{X}_j\left(x_j\right)$, analytically can be cumbersome. One common approach is to lump the geometric shapes of the ${ego}_i$ and ${ov}_j$ into a combined shape $\mathbb{X}_{ij}\left(x_j\right)$ centered at $x_j$  as the Minkowski sum \cite{Lavalle2006} of the underlying shapes (i.e. $\mathbb{X}_i\oplus\mathbb{X}_j=\mathbb{X}_{ij}$). Then ${ego}_i$ can simply be considered as a point located at $x_i$. See the lower portion of Figure \ref{fig:collision_def}. The collision probability threshold criteria becomes: 
\begin{equation}\label{eq:PrC_combShapea}
\Pr\left(x_i\in\mathbb{X}_{ij}\left(x_j\right)\right)\leq\delta;
\end{equation}
where the probability of collision between ${ego}_i$ and ${ov}_j$ is defined by \cite{Wang2018}:
\begin{equation}\label{eq:PrC_def}
\Pr\left(x_i\in\mathbb{X}_{ij}\mkern-4mu\left(x_j\right)\right) =\mkern-4mu \iint \mkern-6mu I_c\mkern-4mu\left(x_i,x_j\right)p_{x_ix_j}\mkern-4mu\left(x_i,x_j\right)dx_idx_j.
\end{equation}
Here, $p_{x_ix_j}$ is the joint probability density function (PDF) of the random variables $x_i$ and $x_j$ for the position states, and  $I_c \left(x_i,x_j\right)$ is a collision indicator function defined as:
\begin{equation}\label{eq:Ic_def}
I_c\left(x_i,x_j\right) =\begin{cases}
1& \text{if } x_i\in\mathbb{X}_{ij}\left(x_j\right),\\
0& \text{otherwise}.
\end{cases}
\end{equation}
Now using the definition of the joint PDF, we can rewrite $p_{x_ix_j }\left(x_i,x_j\right)=p_{x_i}\left(x_i|x_j\right) p_{x_j}\left(x_j\right)$. This, with the definition of $I_c$, allows us to replace the inner indefinite integral of \eqref{eq:PrC_def} with a definite integral over the shape of $\mathbb{X}_{ij}\left(x_j\right)$ to obtain the general formulation for the probability of collision: 
\begin{equation}\label{eq:PrC_conditional}
\Pr\left(x_i\in\mathbb{X}_{ij}\left(x_j\right)\right)=\int\left[\int_{x_i\in\mathbb{X}_{ij}\left(x_j\right)}p_{x_i}\left(x_i|x_j\right)dx_i\right]p_{x_j}\left(x_j\right)dx_j.
\end{equation}

\subsection{Approximation of the Probability of Collision for Gaussian Uncertainty and Car-like Rectangular Geometries}\label{subsec:carlike}

The above general form in \eqref{eq:PrC_conditional} is typically quite difficult to evaluate efficiently for vehicles of non-negligible sizes and arbitrary shapes using on-board computing resources.  To expedite computation of the probability of collision online some simplifying assumptions need to be invoked. We make two assumptions here:

\textbf{\textit{Assumption 1}:} OVs are rectangular in shape and their position states follow Gaussian distributions. This assumption aligns with common practical autonomous vehicle guidance schemes that use lidar- or vision-based perception techniques with bounding boxes for object detection and tracking \cite{Petrovskaya2009,Sivaraman2013} along with Gaussian noise assumptions. Following these practices:
\begin{itemize}
\item Both ${ego}_i$ and ${ov}_j$ are described as car-like rectangular shapes;
\item The PDFs of the position $x$ and heading $\phi$ states of ${ego}_i$ and ${ov}_j$ are independent and available from the respective estimation routines as Gaussians.
\end{itemize}
We designate the normal PDF with $\mathcal{N}\left(\mu,\Sigma\right)$, where $\mu$ is the mean and $\Sigma$ the covariance. In the discussion that follows, we use the state variable in the subscript to differentiate which mean or covariance is intended. We note that the assumption of rectangular shapes is not restrictive, as other geometries may be bounded by a rectangle or divided into multiple sub-regions, each of which is bounded by a rectangle. Further, the motions of the vehicles may have a coupled (interactive) dynamics in traffic for which interaction-aware planning methods have been proposed elsewhere \cite{Sefati2017,Rodrigues2018}. While this is not the focus of the present paper, we note that one could estimate the uncertainty in the joint states/behaviors of $ego_i$ and $ov_j$ as in \cite{Zhou2018} and extend the decoupling approach detailed in this paper for bivariate distributions to higher dimensions and proceed to utilize the proposed constraint tightening framework. This extension is left as a possible future research direction.

\begin{figure}[!t]
\centering
\includegraphics[width=3.25in]{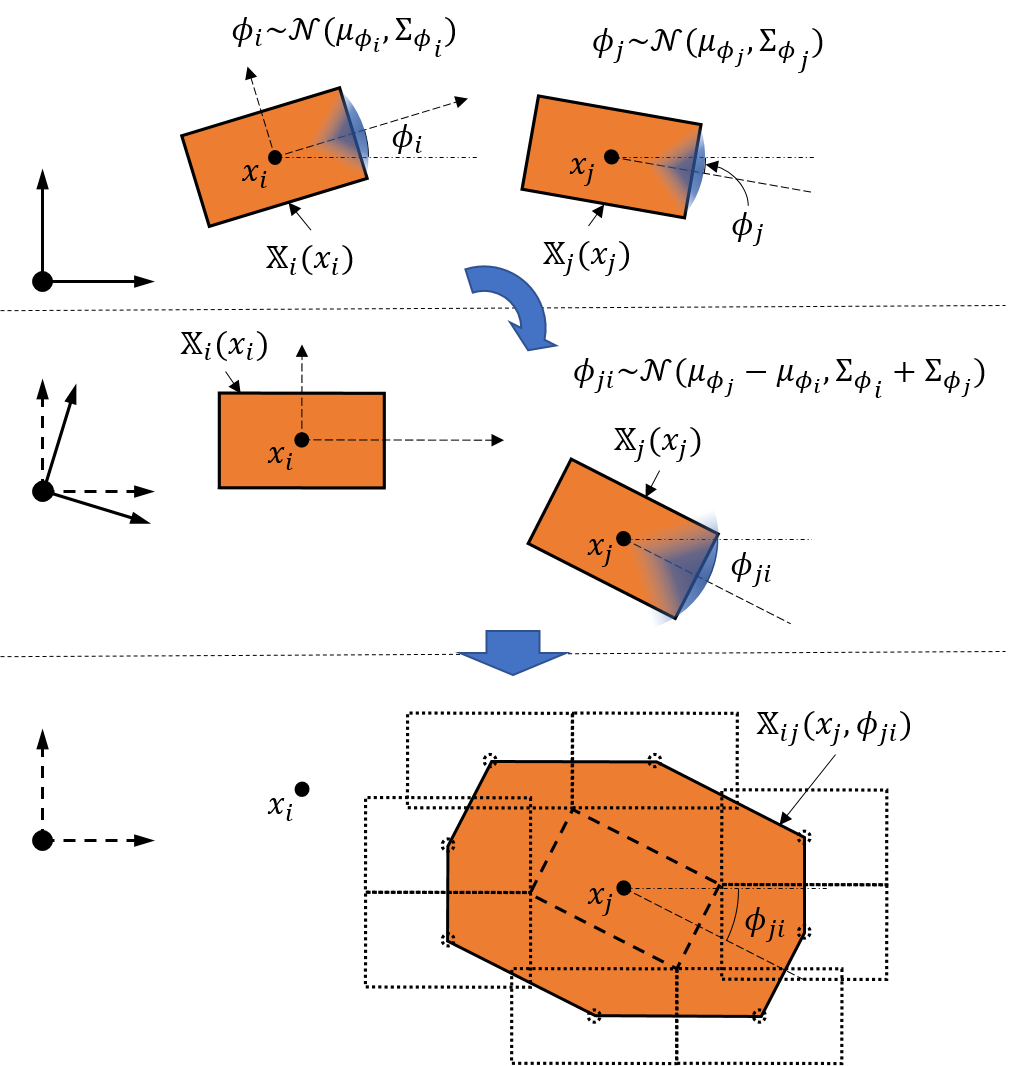}
\caption{Depiction of heading angles $\phi_i$ and $\phi_j$ (top), the rotation to align the reference frame with ${ego}_i$ and the resulting relative heading angle $\phi_{ji}$ (middle), and the combined shape $\mathbb{X}_{ij}\left(x_j,\phi_{ji}\right)$ (bottom).}
\label{fig:rel_angle}
\end{figure}
When we consider both ${ego}_i$ and ${ov}_j$ to be rectangular, instead of mere points or circles, it is necessary to consider the orientation relative to the coordinate frame as well. We can incorporate heading information in our modeling with the following approach. Suppose $\phi_i$ and $\phi_j$ are the respective heading angles of the two vehicles relative to a road fixed frame (see top of Figure \ref{fig:rel_angle}). We rotate the frame of reference to align with the ego vehicle and define the relative heading angle $\phi_{ji}=\phi_j-\phi_i$ with mean $\mu_{\phi_j}-\mu_{\phi_i}$ and covariance $\Sigma_{\phi_j}+\Sigma_{\phi_i}$ (see middle of Figure \ref{fig:rel_angle}). Let $\mathbb{X}_{ij}\left(x_j,\phi_{ji}\right)$ be the combined geometric description located at the 2D position state $x_j$ and constructed according to the relative orientation $\phi_{ji}$ (see bottom of Figure \ref{fig:rel_angle}). Applying Baye’s rule, the probability of collision can be formulated as:
\begin{equation}\label{eq:PrC_wHeading}
\Pr\left(x_i\in\mathbb{X}_{ij}\left(x_j\right)\right)=\int p_{\phi_{ji}}\left(\phi_{ji}\right)\Pr\left(x_i\in\mathbb{X}_{ij}\left(x_j,\phi_{ji}\right)|\phi_{ji}\right)d\phi_{ji}.
\end{equation}

\textbf{\textit{Proposition 1}:} Given independent Gaussians $x_i$ and $x_j$, the probability of collision conditioned on relative heading angle $\phi_{ji}$ is:
\begin{equation}\label{eq:PrC_givenPhi}
\Pr\left(x_i\in\mathbb{X}_{ij}\left(x_j,\phi_{ji}\right)|\phi_{ji}\right)=\mkern-32mu \int\limits_{x_{ij}\in\mathbb{X}_{ij}\left(0,\phi_{ji}\right)} \mkern-40mu  \mathcal{N}_{x_{ij}}\left(\mu_{x_{ij}},\Sigma_{x_{ij}}\right)dx_{ij},
\end{equation}
where $x_{ij}=x_i-x_j$ is the relative position state variable with mean $\mu_{x_{ij}}=\mu_{x_i}-\mu_{x_j}$, and covariance $\Sigma_{x_{ij}}=\Sigma_{x_i}+\Sigma_{x_j}$. 

\textbf{\textit{Proof}:} Proposition 1 was originally proved in \cite{Wang2018}. \hfill$\blacksquare$

When considering heading deviations, it is still cumbersome to compute the collision probability defined by \eqref{eq:PrC_wHeading} and \eqref{eq:PrC_givenPhi} (for real time use), due to the double integral and the fact that the integration bounds in \eqref{eq:PrC_givenPhi} depend on $\phi_{ji}$ and are irregular. This leads us to our next pragmatic assumption.

\textbf{\textit{Assumption 2}:} We use conservative embedding of ${ov}_j$ within a circumscribed rectangle aligned with the axes of ${ego}_i$ considering a specified range of the relative heading angle $\phi_{ji}$.

This assumption is utilized in such a way that the combined geometric description $\mathbb{X}_{ij}\left(0,\phi_{ji}\right)$ remains rectangular, thereby simplifying the evaluation of the integral in \eqref{eq:PrC_givenPhi}. There are, however, two remaining difficulties: first is the consideration of uncertainty in relative heading angle, and second is the coupling of the uncertainty in the two dimensions of the position space.
 
\begin{figure}[!t]
\centering
\includegraphics[width=3.25in]{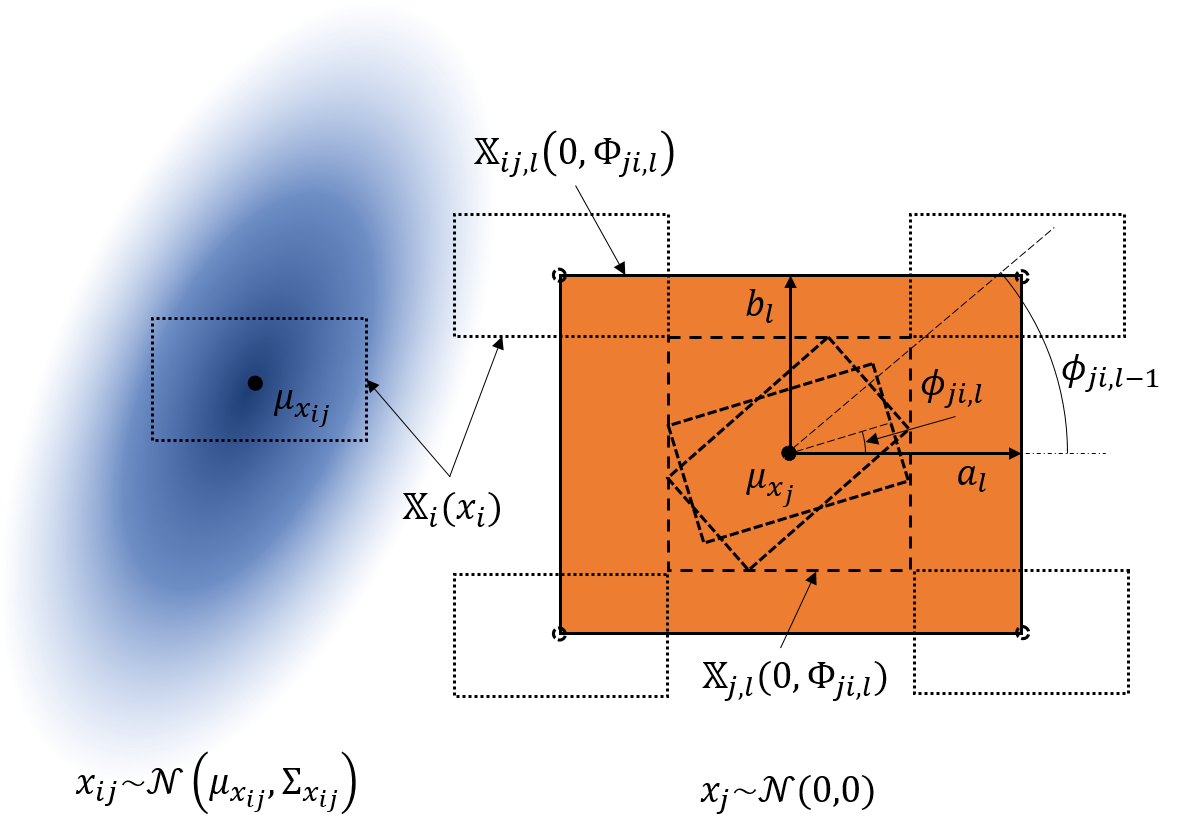}
\caption{Visualization of the conservative embedding of ${ov}_j$ within an ${ego}_i$ axes-aligned circumscribing rectangle for $\phi_{ji}$ interval $l$.}
\label{fig:cons_embed}
\end{figure}

To address the first difficulty, we adopt a method proposed in \cite{Hardy2013}. We consider the support of the relative heading $\phi_{ji}$ and discretize it into $n_\phi$ intervals $\Phi_{ji,l}=\left[\phi_{ji,l-1},\phi_{ji,l}\right], l\in\left\{1,…,n_\phi\right\}$, from $\phi_{ji,0}=\mu_{\phi_{ji}}-\pi/2$ to $\phi_{ji,n_\phi}=\mu_{\phi_{ji}}+\pi/2$. As $\phi_{ji}$ is Gaussian, and by definition has an infinite support, we define $\Phi_{ji,0}=\left[-\infty,\phi_{ji,0}\right]$ and $\Phi_{ji,n_\phi+1}=\left[\phi_{ji,n_\phi},\infty\right]$, thus $\sum_{l=0}^{n_\phi+1}\Pr \left(\phi_{ji}\in\Phi_{ji,l}\right)=1$. For each heading angle interval $\Phi_{ji,l}, l\in\{0,…,$ $n_\phi + 1\}$, the combined shape $\mathbb{X}_{ij}\left(x_j,\Phi_{ji,l}\right)$ is constructed by a Minkowski sum of the shape of ${ego}_i$ and the rectangle that circumscribes the convex hull of the shape of ${ov}_j$ over the interval (see Figure \ref{fig:cons_embed}). We denote the half-length and half-width of $\mathbb{X}_{ij}\left(0,\Phi_{ji,l}\right)$ by $a_l$ and $b_l$, respectively.

An upper bound on the probability of collision, considering the $n_\phi$ discrete intervals of $\phi_{ji}$, can then be obtained as:
\begin{equation}\label{eq:PrC_ub}
\Pr\left(x_i\in\mathbb{X}_{ij}\left(x_j\right)\right)\leq\sum_{l=0}^{n_\phi+1}\Pr\left(\phi_{ji}\in\Phi_{ji,l}\right)\Pr\left(x_i\in\mathbb{X}_{ij}\left(x_j,\Phi_{ji,l}\right)|\Phi_{ji,l}\right).
\end{equation}
The upper bound in \eqref{eq:PrC_ub} is significantly easier to evaluate than \eqref{eq:PrC_wHeading} as the coupled integration over $\phi_{ji}$ and the 2D deviation state $x_{ij}$ is replaced by a (probability) weighted summation of sub-integrations over only the 2D position states.

The second difficulty remains as the collision probability given a specific relative heading angle interval still requires bivariate integration over the 2D position states. To overcome this, we seek a transformation of the deviation state variable into a reference frame that decouples/factorizes the integration of the bivariate Gaussian distribution into univariate integrals. Specifically, we are interested in obtaining a transformation applied to the deviation variables that results in a diagonalized covariance matrix.  As such a transformation may result in an irregular integration region $\mathbb{X}_{ij}^{'}\left(0,\Phi_{ji,l}\right)$, we will also determine the rectangular region $\mathbb{X}_{ij}^{''}\left(0,\Phi_{ji,l}\right)$ that conservatively embeds $\mathbb{X}_{ij}^{'}\left(0,\Phi_{ji,l}\right)$. We explore two methods that accomplish this, unitary scaling (US) and principal axes rotation (PA).

\subsubsection{Decoupling Method 1 -- Unitary Scaling}\label{subsubsec:US}
\begin{figure}[!t]
\centering
\includegraphics[width=5in]{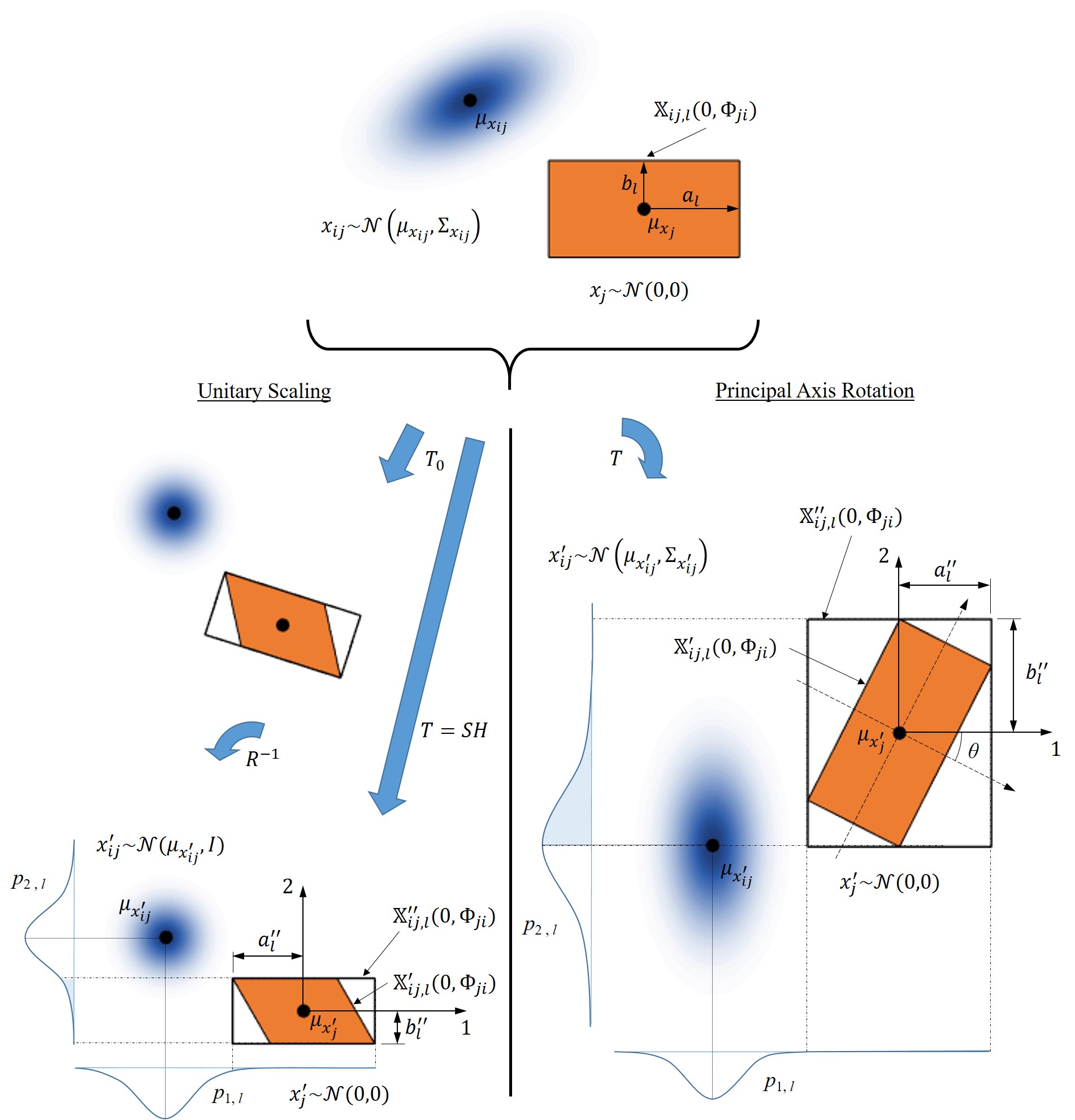}
\caption{Visualization of the decoupling methods, unitary scaling (left), principal axis rotation (right).}
\label{fig:decoupling}
\end{figure}

US normalizes the covariance of the deviation state (i.e. $\Sigma_{x_{ij}^{'}}=T\Sigma_{x_{ij}}T^T=I$, where $I$ is the 2 by 2 identity matrix), in order to allow decoupling in any direction \cite{Hardy2013,Paielli1997}. One transformation matrix that accomplishes this is given by $T_0=\Sigma_{x_{ij}}^{-1/2}$ (see \ref{app:T0} for a discussion on how to compute $T_0$). The transformation $T_0$, as depicted in the middle left hand portion of Figure \ref{fig:decoupling}, can be shown to be composed from three linear elementary operations: rotation, scaling and shear; or $T_0=RSH$, where $R$ is a rotation matrix, $S$ a scaling matrix, and $H$ a shear matrix (see more below). Scaling does not alter the shape, i.e. it maintains a rectangular integration region. Rotation rotates the rectangular integration region with respect to the coordinate axes, however, it does not alter the shape. Taking advantage of the fact that we can decouple $x_{ij}^{'}$ in any direction, we can undo the rotation using the transformation $R^{-1}$ to align $\mathbb{X}_{ij}^{'}\left(0,\Phi_{ji,l}\right)$ with the coordinate axes. See the bottom left of Figure \ref{fig:decoupling}. (With a slight abuse of notation, we designate the axes of the new reference frame by 1 and 2, however, it should be noted that this is not the same frame as the principle axes 1 and 2 mentioned later for Method 2.) Conversely, shear does alter the shape of $\mathbb{X}_{ij}^{'}\left(0,\Phi_{ji,l}\right)$. As shear maintains parallelism, the (tightest) bounding box $\mathbb{X}_{ij}^{''}\left(0,\Phi_{ji,l}\right)$ in the transformed coordinate frame contains either: the two edges along the length of the transformed combined shape $\mathbb{X}_{ij}^{'}\left(0,\Phi_{ji,l}\right)$, we will call this case 1 (shear occurs along axis 1 – depicted on the left of Figure \ref{fig:decoupling});  or the two edges along the width, which we call case 2 (shear occurs along axis 2 – not depicted).

To efficiently implement US, we can reduce the number of matrix operations by taking advantage of the decomposition of $T_0$ into $RSH$ (see \ref{app:decompT0} for how to compute $R$, $S$, and $H$ for both case 1 and case 2). Instead of applying the transformation $T_0$ and  rotating the reference frame by $R^{-1}$, we can simply apply the transformation $T=SH$, which takes the form:
\begin{equation}\label{eq:SH}
SH= \begin{cases}\bigg[\begin{matrix}s_1 & s_1h\\
0& s_2\end{matrix}\bigg] & \text{for case 1,}\\
\bigg[\begin{matrix}s_1&0\\
s_2h & s_2\end{matrix}\bigg]&\text{for case 2}.
\end{cases}
\end{equation}
Based on the assumed forms of $R$, $S$, and $H$, it is possible to solve for $s_1$, $s_2$, and $h$ as functions of the eigenvalues $\lambda_c$, $c\in\left\{1,2\right\}$, and components of the first eigenvector $v_{1c}$, $c\in\left\{1,2\right\}$  of the original covariance matrix $\Sigma_{x_{ij}}$. The complete derivation for $s_1$, $s_2$, and $h$ may be found in \ref{subapp:case1} for case 1 and \ref{subapp:case2} for case 2, while the relationships between the elements of a covariance matrix $\Sigma$ and its eigenvalues and eigenvectors may be found in \ref{app:PAcalcs}.

\subsubsection{Decoupling Method 2 -- Principal Axes Rotation}\label{subsubsec:PA}
The second method attempts to decouple the bivariate Gaussian distribution of the deviation state by a rotation to align the coordinate axes with the principal axes 1 and 2 of the uncertainty. The transformation is then $T= \begin{bmatrix}\cos\theta&-\sin\theta\\\sin\theta&\cos\theta\end{bmatrix}$, where the angle $\theta$ to the principal axes is the direction of an eigenvector of the covariance matrix $\Sigma_{x_{ij}}$ \cite{Constable2008}. The transformed covariance is then $\Sigma_{x_{ij}^{'}}=T\Sigma_{x_{ij}}T^T=D$, where $D$ is the diagonal matrix composed of the eigenvalues of the original covariance matrix $\Sigma_{x_{ij}}$. Let $\theta$ correspond to the eigenvector associated with the first eigenvalue along the diagonal of $D$. The reader is directed to \ref{app:PAcalcs} or our paper \cite{Wang2017a} for detailed equations for this method. The right half of Figure \ref{fig:decoupling} depicts this rotation.

After the transformation for either decoupling method, a rectangle $\mathbb{X}_{ij}^{''}\left(0,\Phi_{ji,l}\right)$ aligned with axes 1 and 2 is found that circumscribes the transformed integration region $\mathbb{X}_{ij}^{'}\left(0,\Phi_{ji,l}\right)$ (depicted for both methods in Figure \ref{fig:decoupling}). We can then solve for the half-length and half-width of the bounding box of the transformed shape, $a_l^{''}$ and $b_l^{''}$, respectively, with the following relationship:
\begin{equation}\label{eq:abdp}
\begin{bmatrix}a_l^{''}\\b_l^{''}\end{bmatrix}=\text{abs}\left(T\right)\begin{bmatrix}a_l\\b_l\end{bmatrix},
\end{equation}
where we use $\text{abs}\left(\cdot\right)$ to describe the elementwise absolute value function. Note that $T$ is the rotation matrix for PA and $T=SH$ for US. The bivariate integral in \eqref{eq:PrC_givenPhi} becomes a product of two univariate integrals:
\begin{equation}\label{eq:PrC_givenPhi_ub}
\Pr\left(x_i\in\mathbb{X}_{ij}\left(0,\Phi_{ji}\right)|\phi_{ji,l}\right)\leq\prod_{c=1}^{2}\int_{\underline{x}_{c,l}^{'}}^{\overline{x}_{c,l}^{'}}\mathcal{N}_{x_c^{'}}\left(\mu_{x_c^{'}},\sigma_{x_c^{'}}^2\right)dx_c^{'},
\end{equation}
where $x_{ij}^{'}=\left[x_1^{'},x_2^{'}\right]^T$  is the transformed relative position state, and $\overline{x}_{c,l}^{'}$  and $\underline{x}_{c,l}^{'}$ are the upper and lower bounds of the conservative transformed shape $\mathbb{X}_{ij,l}^{''}\left(0,\Phi_{ji,l}\right)$ in the c-direction, $c\in\left\{1,2\right\}$ ($\pm a_l^{''}$ along axis 1 and $\pm b_l^{''}$ along axis 2). For future use, we will denote the integral (or the factors) on the right-hand side of \eqref{eq:PrC_givenPhi_ub} as $\Pr\left(x_c^{'}\in\left[\underline{x}_{c,l}^{'},\overline{x}_{c,l}^{'}\right]\right)$. Using the result in \eqref{eq:PrC_givenPhi_ub} and the definition of the cumulative distribution function (CDF) of the normal distribution, we can update the upper bound in \eqref{eq:PrC_ub} as:
\begin{equation}\label{eq:PrC_ubAnalytic}
\Pr\left(x_{ij}^{'}\in\mathbb{X}_{ij}^{''}\left(0\right)\right)= \sum_{l=0}^{n_\phi+1}\left[\Pr\left(\phi_{ji}\in\Phi_{ji,l}\right)\prod_{c=1}^{2}\Psi\left(\frac{x_c^{'}-\mu_{x_c^{'}}}{\sigma_{x_c^{'}}}\right)\Bigg|_{\underline{x}_{c,l}^{'}}^{\overline{x}_{c,l}^{'}}\right],
\end{equation}
where $\Psi\left(z\right)$ is the CDF of the standard normal and $\sigma_{x_c^{'}}$ is the standard deviation. With \eqref{eq:PrC_ubAnalytic}, we have now derived an analytical solution for an upper bound on the mutual probability of collision between ${ego}_i$ and ${ov}_j$.

\subsection{Comparison of Decoupling Methods}\label{subsec:decoupling_comp}

\begin{figure}[!t]
\centering
\includegraphics[width=3.5in]{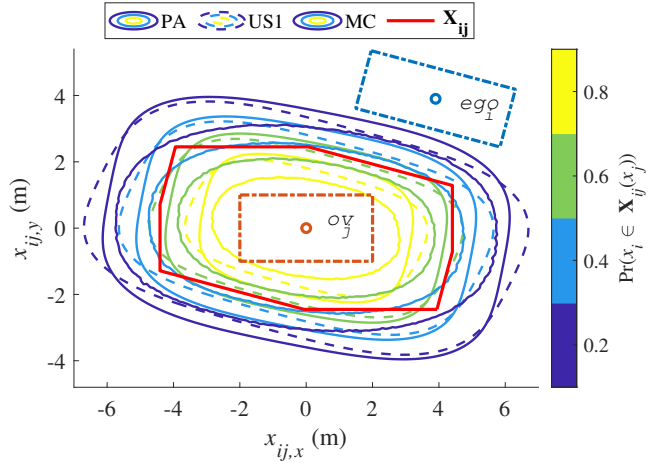}
\caption{Contour of collision probabilities between ${ego}_i$ and ${ov}_j$ using unitary scaling case 1 (US1), principle axes rotation (PA), and Monte-Carlo simulations (MC); $n_{MS}=10,000$ samples were used for the MS method; for the approximation methods $n_\phi=20$ was used.}
\label{fig:prob_contours}
\end{figure}

We now investigate the shape contours for different (constant) probabilities of collision approximated with the two decoupling methods. To obtain the true probability of collision for comparison purposes, we utilize Monte-Carlo Simulations. Specifically, we estimate the true probability of collision by sampling $n_{MS}$ instances of the position and heading states of the ${ego}_i$ and ${ov}_j$ (both with mean $\mu_{x,\phi}$ and joint covariance $\Sigma_{x,\phi}$) and then calculating the indicator function \eqref{eq:Ic_def} and the probability of collision by:
\begin{equation}
\Pr\left(x_i\in\mathbb{X}_{ij}\right)=\frac{1}{n_{MS}} \sum_{k=0}^{n_{MS}}I_c\left(\begin{bmatrix}x_{i,k}\\\phi_{i,k}\end{bmatrix},\begin{bmatrix}x_{j,k}\\\phi_{j,k}\end{bmatrix}\right)
\end{equation}
Figure \ref{fig:prob_contours} shows the collision probability contours approximated using the unitary scaling case 1 (US1), principal axes (PA) rotation, and Monte-Carlo Simulations (MS) methods. We see that both approximation methods (US1, PA) are conservative as the true probability contours are contained within the contours of the approximations.

In the following, we seek to characterize the level of conservatism for both decoupling methods. This can be very important to consider for online motion planning tasks, as the choice of decoupling method could mean the difference between finding a feasible path in the available computing time or not. Substituting the respective definitions of $T$ for US and PA into \eqref{eq:abdp} yields:
\begin{equation}\label{eq:abdp_cases}
\begin{bmatrix}a^{''}\\
b^{''}\end{bmatrix} = \begin{cases}
\bigg[\begin{matrix}\left|s_1\right|a+\left|s_1h\right|b\\
\left|s_2\right|b\end{matrix}\bigg] & \text{for US case 1,}\\
\bigg[\begin{matrix}\left|s_1\right|a\\
\left|s_2h\right|a+\left|s_2\right|b\end{matrix}\bigg] & \text{for US case 2,}\\
\bigg[\begin{matrix}\left|\cos\theta\right|a+\left|\sin\theta\right|b\\
\left|\sin\theta\right|a+\left|\cos\theta\right|b\end{matrix}\bigg] & \text{for PA}.
\end{cases}
\end{equation}
To compare the two decoupling methods, we emphasize that both cases of US scale the original dimensions, whereas PA does not. As scaling does not result in any conservatism, we will compare $a^{''}$ and $b^{''}$ of PA to $a^{''}/\left|s_1\right|$ and $b^{''}/\left|s_2\right|$ of either US case. We do this both analytically and numerically.

First for PA, it is easy to show geometrically, or by taking derivatives of \eqref{eq:abdp_cases} with respect to $\theta$, that both $a^{''}$ and $b^{''}$ are bounded in the set $\left[\min\left\{a,b\right\},\sqrt{a^2+b^2}\right]$. If $b<a$, then the minimum of $a^{''}$ occurs at the angles $\theta\in\left\{\pi/2\pm k\pi\right\}$ for all $k\in\mathbb{I}_{\geq0}$, and the minimum of $b^{''}$ occurs at $\theta\in\left\{\pm k\pi\right\}$ for all $k\in\mathbb{I}_{\geq0}$, where $\mathbb{I}_{\geq0}$ is the set of positive integers including 0. If $a>b$, then the angles obtaining the minimum of $a^{''}$ and $b^{''}$ are each shifted by $\pi/2$. The maximum of $a^{''}$ occurs at $\theta\in\left\{\tan^{-1}\left(b/a\right)\pm k\pi\right\}$ for all $k\in\mathbb{I}_{\geq0}$, while the maximum of $b^{''}$ occurs at $\theta\in\left\{\tan^{-1}\left(a/b\right)\pm k\pi\right\}$ for all $k\in\mathbb{I}_{\geq0}$. Further, if we look at the area of the transformed integration region for PA, $a^{''}b^{''}$, we can see that it is bounded in $\left[ab,0.5\left(a^2+b^2\right)+ab\right]$, where the minimum area occurs at the angles $\theta\in\left\{\pm k\pi/2\right\}$ for all $k\in\mathbb{I}_{\geq0}$, and the maximum area occurs at the angle $\theta\in\left\{\pi/4\pm k\pi/2\right\}$ for all $k\in\mathbb{I}_{\geq0}$.

\begin{figure*}[!t]
\centering
\includegraphics[width=5.5in]{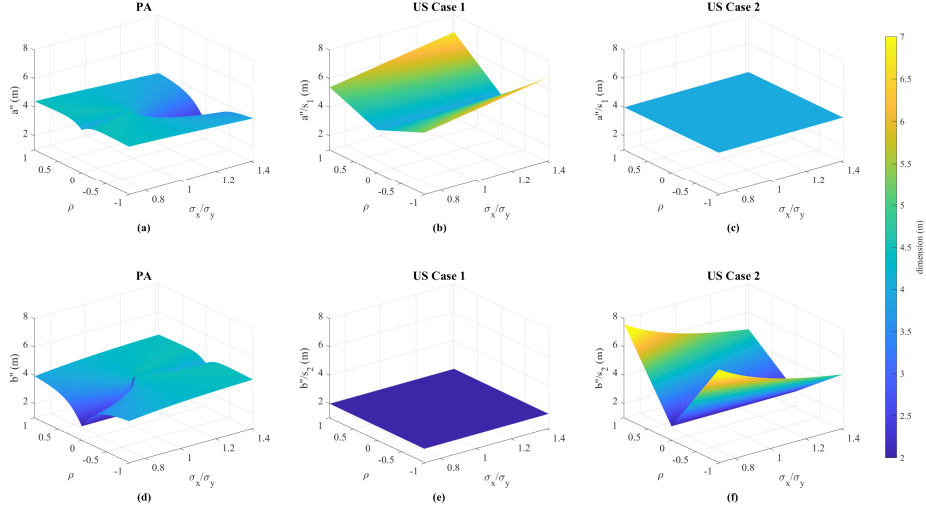}
\caption{Plot of $a^{''}$ (top row) and $b^{''}$ (bottom row), as a function of correlation coefficient $\rho$ and $\sigma_x/\sigma_y$ for principal axis rotation (left), unitary scaling case 1 (center), and unitary scaling case 2 (right), with original dimension $a=4$ and $b=2$.}
\label{fig:conservatism}
\end{figure*}
Now investigating US, it is easy to see from \eqref{eq:abdp_cases} that $b^{''}/\left|s_2\right|=b$ for case 1 and $a^{''}/\left|s_1\right|=a$ for case 2, i.e. there is no conservatism in the respective directions. Therefore, all of the conservatism can be lumped into $a^{''}/\left|s_1\right|$ for case 1 and $b^{''}/\left|s_2\right|$ for case 2. Specifically, the conservatism is proportional to the shear $h$, therefore we seek to analyze how $h$ (case 1 \eqref{eq:hOfvlambdaCase1} and case 2 \eqref{eq:hOfvlambdaCase2} in \ref{app:decompT0}) behaves with respect to $\lambda_1$, $\lambda_2$, $v_{11}$, and $v_{12}$. As we chose unit eigenvectors, we know $v_{11},v_{12}\in\left[0,1\right]$ and $v_{11}^2+v_{12}^2=1$. The denominator of $h$, for either case,  is then a weighted combination of the eigenvalues, while the numerator is bounded with respect to $v_{11}$ and $v_{12}$. Therefore, it is of interest to understand how $h$ behaves with respect to the eigenvalues $\lambda_1$ and $\lambda_2$. In theory, if the eigenvalues tend towards infinity, it can be shown that $h$ tends towards either $\pm v_{12}/v_{11}$ or $\mp v_{11}/v_{12}$ for either case (see \ref{app:decompT0}). As both are unbounded this means the conservatism of US is theoretically unbounded in one direction. However, in practice, the spectral radius of $\Sigma_{x_{ij}}$ is (or the eigenvalues are) typically bounded meaning practical bounds on $h$ can be computed. 

Recalling that the eigenvalues and eigenvectors are a function of the standard deviations $\sigma_x$ and $\sigma_y$, as well as the correlation coefficient $\rho$, we present plots of the dimensions of $\mathbb{X}_{ij}^{''}$ as a function of $\rho$ and $\sigma_x/\sigma_y$ in Figure \ref{fig:conservatism}. As presented analytically, Figure \ref{fig:conservatism} (a) and (d) support that $a^{''}$ and $b^{''}$ for PA are bounded, and therefore the corresponding conservatism is bounded. Then referencing Figure \ref{fig:conservatism} (b) and (e), we see how $b^{''}/\left|s_2\right|$ remains constant while $a^{''}/\left|s_1\right|$ increases, apparently unbounded, as $\left|\rho\right|$ and $\sigma_x/\sigma_y$ increase for US case 1. Similar behavior is observed for the opposite dimensions in Figure \ref{fig:conservatism} (c) and (f) for US case 2.

The key observations are the following: in unstructured environments, where ${ego}_i$ may need to navigate around an obstacle in any direction, the PA decoupling method is beneficial as the conservatism is easily bounded in any direction. However, in applications where only one direction is of relative importance, US guarantees minimal conservatism in one or the other direction. For example, if the application is autonomous driving on a multi-lane highway, US case 1 will guarantee minimal lateral conservatism allowing ${ego}_i$ to pass (or be passed by) OVs in adjacent lanes. On the other hand, on a single lane highly populated road where the ${ego}_i$ is following an OV using adaptive cruise control (ACC), US case 2 is beneficial as it allows for minimal conservatism in the longitudinal direction, and therefore accommodate longitudinally dense traffic.

\section{Application in Closed-Loop Trajectory Planning}\label{sec:closed_loop}

\subsection{Nonlinear MPC-based Motion Planning Formulation}\label{subsec:MPC}

To demonstrate the use of the above constraint tightening methods, we formulate a chance constrained nonlinear MPC (CC-MPC) that solves the trajectory planning problem:
\begin{subequations}\label{eq:ocp}
\begin{gather}
\min_{\mathbf{x}_i,\mathbf{u}}\mathbb{E}\left[\left(\mathbf{y}-\mathbf{r}\right)^T\mathbf{Q}\left(\mathbf{y}-\mathbf{r}\right) + \mathbf{u}^T\mathbf{Ru}\right]\label{eq:cost}
\intertext{Subject to:}
\dot{x}_i=f\left(x_i,u,w\right),x_i\in\mathcal{X},u\in\mathcal{U},w\in\mathcal{W}\label{eq:dynamics}\\
y_i=Cx_i+\nu,\nu\in\mathcal{V}\label{eq:output}\\
x_i\left(0\right)=\hat{x}_i\left(t\right)\label{eq:init_cond}\\
\Pr\left(\mathbb{X}\left(x_i\right)\cap\mathbb{X}\left(x_j\right)\neq\emptyset\right)\leq\delta \ \ \forall j \in \left\{1,...,n_{ov}\right\}\label{eq:collision_const}\\
\mathbb{E}\left[c_2\left(x_i,u\right)\right]\geq0.\label{eq:exp_const}
\end{gather}
\end{subequations}
The problem in \eqref{eq:cost} minimizes the expected error in output tracking, where $\mathbf{y}$, $\mathbf{r}$, and $\mathbf{u}$ are, respectively, the outputs, references to be tracked, and inputs augmented over the prediction horizon. Let $\mathbf{Q}$ and $\mathbf{R}$ be block diagonal matrices composed of the symmetric positive semi-definite weighting matrices $Q_k$ and $R_k$ at each time step $k\in\left\{1,…,n_p\right\}$. The constraints include: 1) the nonlinear motion dynamics model and the feasible state and input bounded sets, $\mathcal{X}$ and $\mathcal{U}$, respectively, in \eqref{eq:dynamics}, with process noise $w$ sampled from the set $\mathcal{W}$; 2) the output model in \eqref{eq:output}, with measurement noise $\nu$ sampled from the set $\mathcal{V}$; 3) the initial condition in \eqref{eq:init_cond}; 4) the probabilistic collision avoidance constraints in \eqref{eq:collision_const}, and 5) nonlinear expectation constraints are modeled by \eqref{eq:exp_const}.

Here the ego vehicle state $x_i=\begin{bmatrix}s&y_e&\phi&v&a&\gamma\end{bmatrix}^T$ includes the position state (longitudinal position $s$ and lateral position $y_e$), heading angle $\phi$, velocity $v$, acceleration $a$, yaw rate $\gamma$, and the inputs include desired acceleration $a_d$ and yaw rate $\gamma_d$. We assume full state measurement, i.e. $C$ is the 6 by 6 identity matrix, however, the optimal control problem only tracks $y_e$ and $v$ references. The kinematics are governed by a particle motion model in the Frenet frame, with the dynamics associated with vehicle mass, moment of inertia, drivetrain, and steering system lumped into first order lag dynamics for $a$ and $\gamma$. For a detailed discussion of the complete nonlinear motion model of \eqref{eq:dynamics} and nonlinear constraints of \eqref{eq:exp_const}, including friction limits, the reader is directed to our prior work in \cite{Weiskircher2017}. We also defer the discussion of the transformation of the stochastic CC-MPC problem above into a deterministic one (with collision constraints tightened with our approach) to references elsewhere \cite{Heirung2018,Lorenzen2017,Kouvaritakis2016}.

We utilize an unscented Kalman filter (UKF) to estimate the mean current state of ${ego}_i$ and the idea of the most likely measurement \cite{DuToit2012} to propagate the covariance over the prediction horizon.
The predicted ${ov}_j$ states include $x_j=\begin{bmatrix}s&y_e&v&\phi\end{bmatrix}^T$. For estimating and predicting $x_j$ and its covariance we use a UKF coupled with the nonlinear motion model from \cite{Weiskircher2017}. 

\subsection{Direct Implementation of the Probabilistic Constraint}\label{subsec:direct}

For a direct implementation of the probabilistic constraint, we replace the general probability on the left-hand side of \eqref{eq:collision_const} with the analytical upper bound on the probability of collision in \eqref{eq:PrC_ubAnalytic}. To simplify calculations within the MPC, we utilize the prior time step plan of $x_i$ and the prediction of $x_j$ to precompute $\Sigma_{x_{ij}}$, $\phi_{ji}$, and $\Sigma_{\phi_{ji}}$. This allows us to calculate the integration bounds $a_l^{''}$ and $b_l^{''}$ for every $l$ relative heading range $\Phi_{ji,l}$ and each time step prior to the MPC call. We solve the optimal control problem (OCP) using the MPCTools \cite{Risbeck2016} interface with CasADi \cite{Andersson2019} and the IPOPT solver \cite{Wachter2006}.

\begin{figure}[!t]
\centering
\includegraphics[width=4.5in]{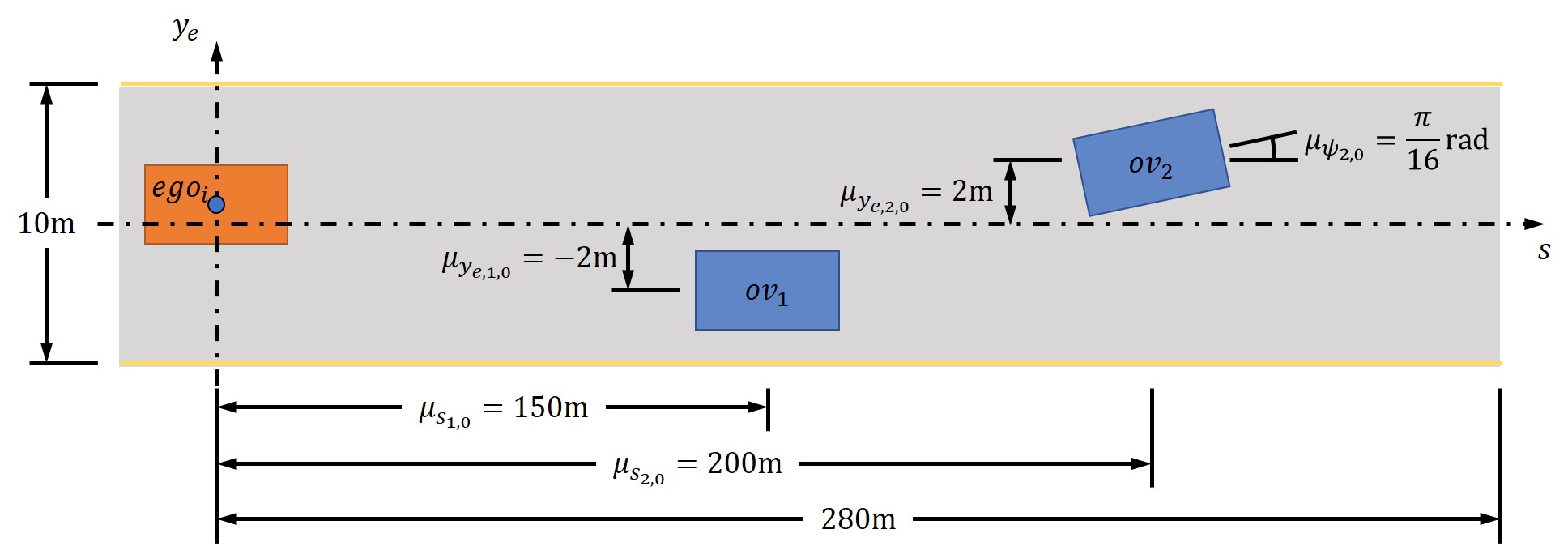}
\caption{Illustrative scenario.}
\label{fig:scenario}
\end{figure}
To illustrate how the constraint tightening framework and CC-MPC perform, we set up an example simulation scenario with an autonomously controlled ego vehicle driving down a road and avoiding two stationary object vehicles within its path. The specific roadway with obstacles is presented in Figure \ref{fig:scenario}. The roadway is 280m long and 10m wide. The ego vehicle ${ego}_i$ starts at an initial position state randomly perturbed about the means $\mu_{s_{i,0}}=0$m, $\mu_{y_{e,i,0}}=0.5$m, and $\mu_{\phi_{i,0}}=0$rad, with variances of $\Sigma_{s_{i,0}}=0.1$m\textsuperscript{2}, $\Sigma_{y_{e,i,0}}=0.1$m\textsuperscript{2}, and $\Sigma_{\phi_{i,0}}=0.01$rad\textsuperscript{2}. Further, ${ego}_i$ has an initial velocity of $v_{i,0}=17$ m/s. The CC-MPC controlling the ego vehicle tracks the lane centerline with $r_{y_e}=0$m and a reference velocity $r_v=20$m/s. The plant model for the $ego_i$ is the same as that used within the CC-MPC. The stationary OVs are then located as follows: ${ov}_1$ has a mean initial position state of $\mu_{s_{1,0}}=150$m, $\mu_{y_{e,1,0}}=-2$m, and $\mu_{\phi_{1,0}}=0$rad; while, ${ov}_2$ has a mean initial position state of $\mu_{s_{2,0}}=200$m, $\mu_{y_{e,2,0}}=2$m, and $\mu_{\phi_{2,0}}=\pi/16$rad. The initial position states of both stationary OVs are randomly perturbed about the means with the same variance as the ${ego}_i$.

\begin{figure}[!t]
\centering
\includegraphics[width=3.5in]{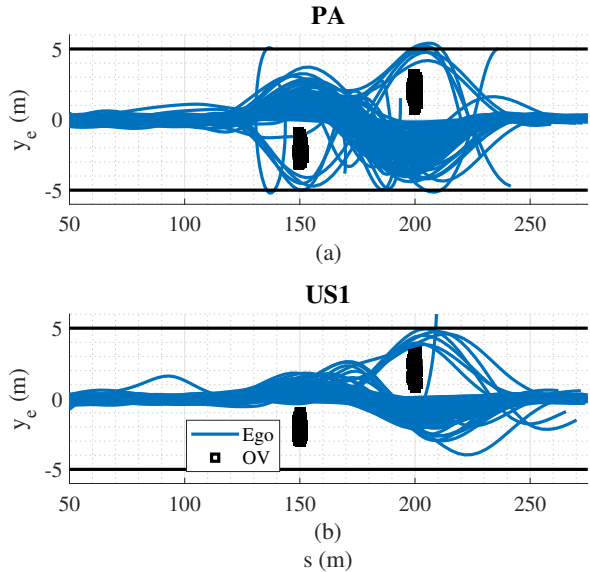}
\caption{Simulation trajectories for direct implementation of the tightened probabilistic collision constraint using (a) PA decoupling and (b) US1 decoupling.}
\label{fig:trajectoriesProb}
\end{figure}

The constraint tightening framework and CC-MPC is updated every 0.15 simulation seconds, and the CC-MPC is discretized with 0.15s time step with a horizon of 6s, for a total of 40 time steps in the horizon. We choose a probability of collision threshold of $\delta=0.001$ and discretize the relative heading space with $n_\phi=20$. For comparison of the two decoupling methods described in Section \ref{subsec:carlike}, 100 simulations each using US1 and PA decoupling methods were completed. The results for the 100 simulations using each decoupling method are plotted within Figure \ref{fig:trajectoriesProb}.

\begin{figure}[!t]
\centering
\includegraphics[width=3.5in]{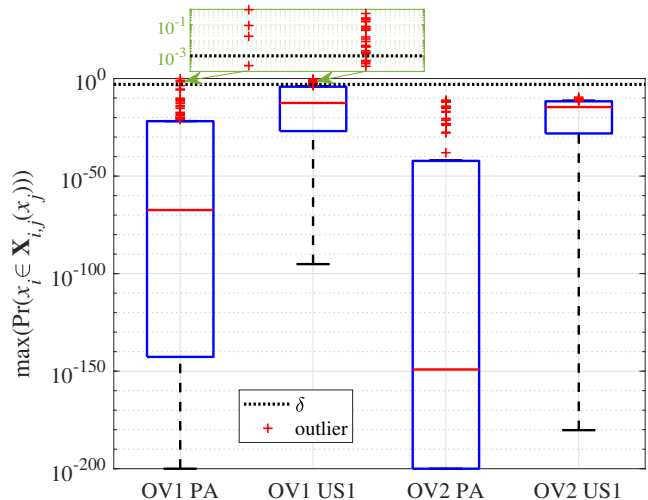}
\caption{Box plot of max observed probability of collision between the ${ego}_i$ and ${ov}_1$ and ${ov}_2$ for each simulation with direct implementation of the tightened probabilistic constraint.}
\label{fig:PrCboxPlotProb}
\end{figure}

From Figure \ref{fig:trajectoriesProb} we can see that the CC-MPC, using either decoupling method, does not always find a solution that keeps the vehicle traveling down the roadway. When comparing the trajectories for PA and US1 decoupling, in Figure \ref{fig:trajectoriesProb} (a) and (b), respectively, we can see that the US1 results in more robust behavior. Specifically, US1 only has one instance where a solution is not found, while PA results in several. Furthermore, with either decoupling method the CC-MPC does not always pass the OVs on the same side. The reason US1 decoupling performs more robustly is because it is less conservative along the relevant (lateral) dimension resulting in additional free space for planning, as discussed earlier in Section \ref{subsec:decoupling_comp}. The statistics of the max probability of collision in Figure \ref{fig:PrCboxPlotProb} support this notion as the medians (red lines) and quartiles (upper and lower extents of the blue box) for US1 decoupling are greater than those for PA. We note that there are outliers which exceed the probability threshold $\delta$ in Figure \ref{fig:PrCboxPlotProb}, as a solution is not always found.

Another drawback is that the implementation of the approximated probabilistic collision avoidance constraint with either method is still computationally expensive. The simulations presented here were run on a modern laptop with an Intel\textregistered \ Core\textsuperscript{TM} i7-7820HQ 2.9 GHz quad core processor with 16.0GB RAM. The median computation time for the CC-MPC with PA decoupling over the 9,300 calls of the solver was 0.142s. This is on the border of the update rate of 0.15s, resulting in the CC-MPC not meeting real-time requirements more than 45\% of the time. This is improved with US1 decoupling, where the median is 0.126s and real-time requirements are not met with 30\% of the CC-MPC calls. To improve the robustness and computational efficiency of our CC-MPC we develop convexified tightened collision avoidance constraints as described below.

\subsection{Convexified Tightened Constraint}\label{subsec:convex_con}

We perform the convexification of the probabilistic collision avoidance constraint in 2 steps: first, we seek a conservative approximation of the probability contour in the position space of ${ego}_i$ for the threshold $\delta$ with each ${ov}_j$; second, we seek convex bounds on the mean position state that guarantee the threshold $\delta$ is not violated.

\subsubsection{Tightened constraint bounding box}\label{subsubsec:bb}

\begin{figure}[!t]
\centering
\includegraphics[width=3.5in]{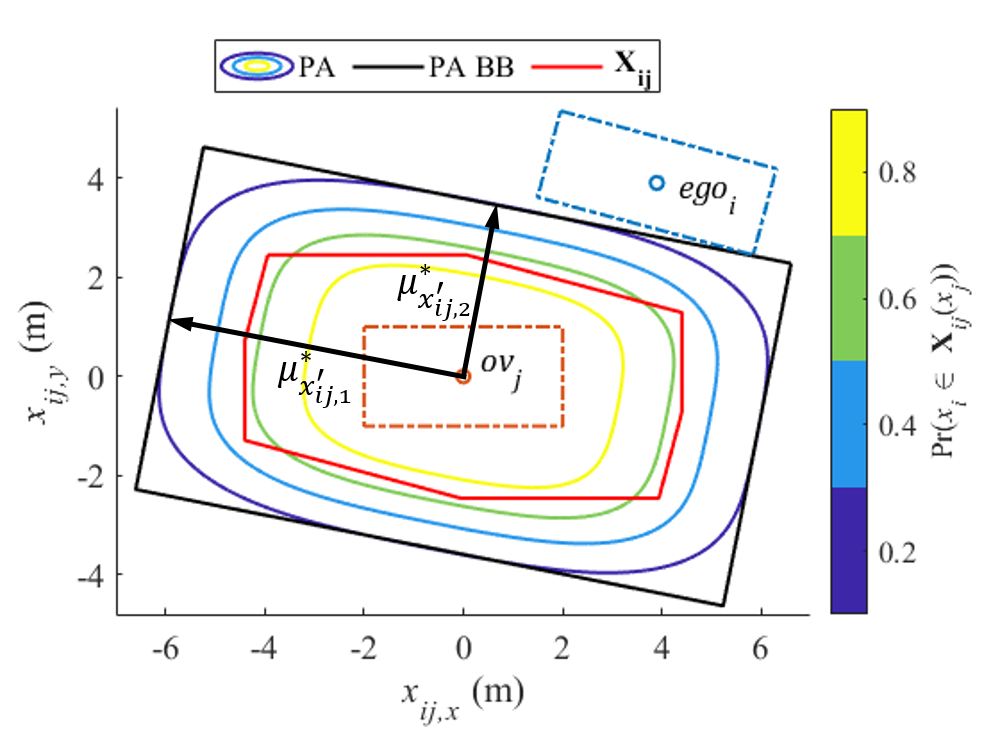}
\caption{Depiction of the tightened constraint bounding box for the principal axis approach. The statistics match those in Figure \ref{fig:prob_contours}.}
\label{fig:prob_contoursBB}
\end{figure}

For a given threshold $\delta$, it remains difficult to invert the right-hand side in \eqref{eq:PrC_ubAnalytic} to find the corresponding contour. We simplify this task by finding the length and width of a bounding box circumscribing the contour (see Figure \ref{fig:prob_contoursBB}) as tightly as possible, thereby remaining conservative but not too much so. To do this, we implement a search along axes aligned with those used in the decoupling approximations.

Now, to find the tightened constraint bounding box we perform five steps:
\begin{enumerate}
\item Discretize the relative heading space into $n_\phi$ intervals $\Phi_{ji,l}$, where $l\in\left\{1,…,n_\phi\right\}$, and calculate $\Pr\left(\phi_{ji}\in\Phi_{ji,l}\right)$, $a_l^{''}$, and $b_l^{''}$ for all $l$.\label{bb_step1}
\item Calculate $\Pr\left(x_1^{'}\in\left[\underline{x}_{1,l}^{'},\overline{x}_{1,l}^{'}\right]\right)$ assuming $\mu_{x_1^{'}}=0$ for all $l\in\left\{1,…,n_\phi\right\}$. \label{bb_step2}
\item Implement a gradient-based probability threshold search (PTS) algorithm presented in Algorithm \ref{alg:PTS} in \ref{app:PTS}, to find $\mu_{x_2^{'}}^*$, the extent along axis 2 where $\Pr\left(x_{ij}^{'}\in\mathbb{X}_{ij}^{''}\left(0\right)\right)=\delta$. \label{bb_step3}
\item Repeat steps 2 and 3 by switching the coordinates (i.e. $\mu_{x_2^{'}}=0$, and search for the corresponding $\mu_{x_1^{'}}^*$ where $\Pr\left(x_{ij}^{'}\in\mathbb{X}_{ij}^{''}\left(0\right)\right)=\delta$). \label{bb_step4}
\end{enumerate}
The resulting tightened constraint bounding box in the transformed coordinate frame is of size $2\mu_{x_1^{'}}^*$ by $2\mu_{x_2^{'}}^*$. We note here that steps \ref{bb_step2} and \ref{bb_step3} in the above iterative algorithm are independent. This provides the opportunity to use different decoupling methods to find $\mu_{x_1^{'}}^*$ and $\mu_{x_2^{'}}^*$. Therefore, to reduce conservatism it is ideal to use US case 1 to find $\mu_{x_2^{'}}^*$ and US case 2 to find $\mu_{x_1^{'}}^*$. To obtain constraints expressed in the original reference frame the inverse transformations for the respective methods may be applied to the points defining the bounding box in the transformed coordinates.

\subsubsection{Convex bounds}\label{subsubsec:conv_bounds}
When directly constraining the mean position state of ${ego}_i$ to be outside the tightened constraint bounding box found above (Figure \ref{fig:prob_contoursBB} for a given threshold $\delta$), a disjunctive or mixed-integer program results requiring specialized solvers. In our prior work, to avoid posing the problem in this way, we utilized elliptical or hyper-elliptical collision avoidance constraints, as they are continuous and differentiable and can be readily handled via nonlinear optimization solvers. There are, however, two drawbacks to that method: 1) the feasible set for the position states $\begin{bmatrix}s&y_e\end{bmatrix}^T$ are non-convex; and, 2) the major and minor axes of a hyper-ellipse (fourth order) must be enlarged by a factor of at least $2^{1/4}$, in order to circumscribe the minimum bounding box \cite{Wang2017b}.

\begin{figure}[!t]
\centering
\includegraphics[width=3.5in]{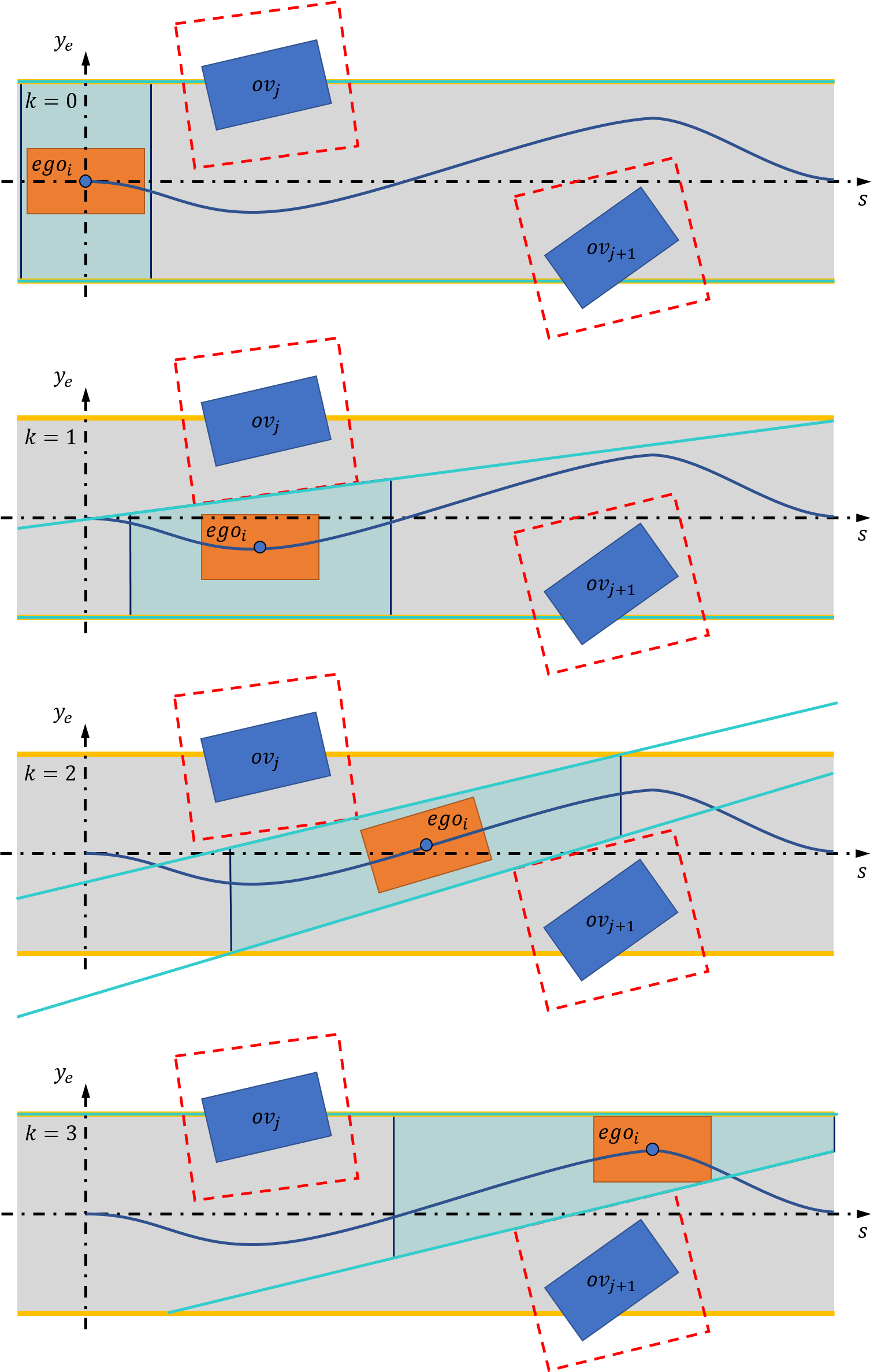}
\caption{Example evolution of the convex constraint space (shaded teal) over the prediction horizon. The blue spline is the predicted trajectory, red dashed lines are the bounding boxes, yellow lines are the road boundary, and the teal lines are the linear $y_e$ bounds.}
\label{fig:convex_con}
\end{figure}
Here, we choose a third approach: at each time step in the horizon, the lateral deviation $y_e$ is upper and lower constrained by linear functions of $s$, as depicted in Figure \ref{fig:convex_con}. More specifically, the upper and lower constraints on $y_e$ are defined as:
\begin{equation}\label{eq:conv_bounds}
\underline{\alpha}s+\underline{\beta}\leq y_e\leq\overline{\alpha}s+\overline{\beta},
\end{equation}
where $\overline{\alpha}$ and $\underline{\alpha}$ are the slopes of the upper and lower bounds, respectively, on $y_e$ at time step $k$, while $\overline{\beta}$ and $\underline{\beta}$ are the respective intercepts. We further constrain the longitudinal position $s$ to be within an upper bound $\overline{s}$ and lower bound $\underline{s}$ to guarantee the constraint \eqref{eq:conv_bounds} performs as intended.

In order to implement this method for deterministic constraint tightening we make two assumptions: 1) the solution to the MPC problem does not deviate significantly from one solver call to another, and 2) if the prior plan is shifted assuming a constant velocity between prediction time steps, as presented in \cite{Goulet2021}, the error in the position states is negligible. The first assumption allows us to use the prior solution of the MPC problem to define the slope and intercept of the boundary line at each time step in the prediction horizon. Whereas the second assumption allows for an efficient method of shifting the plan if the time step in the horizon is larger than the update rate of the MPC.

Prior to defining the slopes and intercepts, we will introduce some useful notations. Let $\mathbb{B}_j$ denote the ordered set of four $\left(s,y_e\right)$ points that define the tightened constraint bounding box of ${ov}_j$. Moving forward we will drop the OV index $j$ to reduce clutter. Each point in $\mathbb{B}$ is designated by $b_p,\ p\in\left\{1,…,4\right\}$. An edge of the tightened constraint bounding box is defined by two adjacent points in $\mathbb{B}$ and designated by $B_p=\left\{b_p,b_q\right\}$, where $q=p+1$ if $p<4$ and $q=1$ if $p=4$.

To find $\overline{\alpha},\ \underline{\alpha},\ \overline{\beta}$, and $\underline{\beta}$ we begin by sorting whether a given ${ov}_j$ should be considered for the upper or lower bound. For the purposes of this paper, if the geometric center of ${ov}_j$ is above the lane centerline ($y_{e,j}>0$), then it is considered in the upper bound on $y_e$. While, if $y_{e,j}$ is below the lane centerline $y_{e,j}\leq0$, it is considered in the lower bound on $y_e$. Once the OVs are sorted, a simple set of rules is used to calculate $\overline{\alpha},\ \underline{\alpha},\ \overline{\beta}$, and $\underline{\beta}$. There are three possible conditions based on $s_i$; we will refer to Figure \ref{fig:convex_con} to describe each condition. We will use the subscript $s$ (or $y_e$) when referring to those respective values of elements in $\mathbb{B}$.  Next we will overview how to calculate $\overline{\alpha}$ and $\overline{\beta}$. An analogous method may be used to calculate $\underline{\alpha}$ and $\underline{\beta}$, however, we will highlight minor changes as necessary.

The first condition arises if $\min\mathbb{B}_s\leq s_i \leq \max\mathbb{B}_s$, as is the case at $k=1$ for the upper $y_e$ bound in Figure \ref{fig:convex_con} (or $k=3$ for the lower). To calculate $\overline{\alpha}$ and $\overline{\beta}$, we begin by determining the set $\left\{B_p\right\}$ of edges where $s_i\in\left[\min B_{p,s}, \max B_{p,s}\right]$. Then $\overline{\alpha}$ and $\overline{\beta}$ are the slope and intercept, respectively, of the edge $B_p$ in that set satisfying $\argmin_p\left\{B_{p,y_e}\right\}$  (or $\argmax_p\left\{B_{p,y_e}\right\}$ for $\underline{\alpha}$ and $\underline{\beta}$).

The second possible condition occurs if ${ego}_i$ is within $\Delta s$, a fixed look-ahead/behind distance, of the tightened constraint bounding box. Specifically, if $s_i\geq\max\mathbb{B}_s$ and $s_i-\Delta s\leq\max\mathbb{B}_s$ or $s_i\leq\min\mathbb{B}_s$ and $s_i+\Delta s\geq\min\mathbb{B}_s$ (both depicted at $k=2$ in Figure \ref{fig:convex_con}, the former as the upper bound and the latter as the lower). We then calculate $\overline{\alpha}$ and $\overline{\beta}$ as follows:
\begin{subequations}\label{eq:alphaBeta}
\begin{gather}
\overline{\alpha}=\frac{\overline{y}_e-b_{p,y_e}}{s_i\mp\Delta s-b_{p,s}}\\
\overline{\beta}=\overline{y}_e-\overline{\alpha}\left(s_i\mp\Delta s\right),
\end{gather}
\end{subequations}
where $\overline{y}_e$ is the upper roadway boundary. For $\underline{\alpha}$ and $\underline{\beta}$ replace $\overline{y}_e$ in \eqref{eq:alphaBeta} with the lower roadway boundary $\underline{y}_e$.

The last possible condition occurs when the tightened constraint bounding box is beyond the look ahead/behind distance, i.e. when $s_i+\Delta s<\min\mathbb{B}_s$ or $s_i-\Delta s>\max\mathbb{B}_s$. When this occurs, the constraint is defined by the roadway boundaries, therefore $\overline{\alpha}=0$ and $\overline{\beta}=\overline{y}_e$ ($\underline{\beta}=\underline{y}_e$). This condition is depicted at time steps $k=0$ or 3 for the upper bound in Figure \ref{fig:convex_con} ($k=0$ and 1 for the lower).

We note that the above algorithm for determining a convex constraint space is relatively simple, and the problem of decomposing the obstacle-free space into convex sets is an intriguing research question on its own with many solutions (e.g. \cite{Brooks1983,Deits2015}). However, our primary focus within this paper is on guaranteeing the probability of collision threshold is not violated (accomplished in Section \ref{subsubsec:bb}) and on the added performance benefits of the convexified tightened constraint.

\subsubsection{Results}
\begin{figure}[!t]
\centering
\includegraphics[width=3.5in]{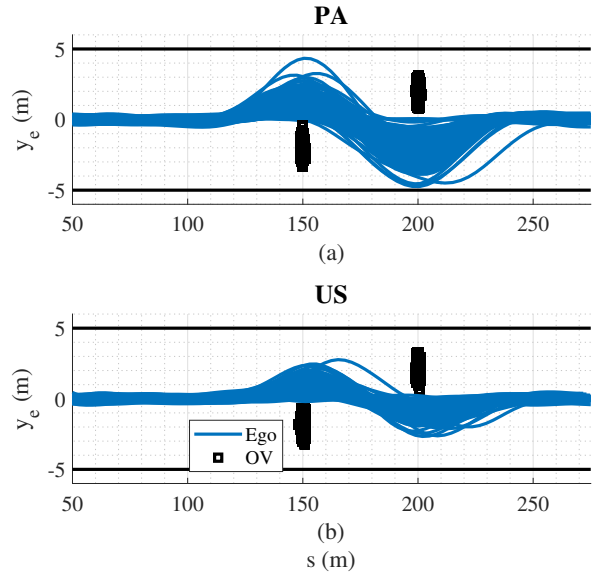}
\caption{Simulation trajectories for CC-MPC with convexified tightened collision constraints using (a) PA decoupling and (b) US decoupling.}
\label{fig:trajectoriesConv}
\end{figure}

In this section, we compare the performance of our CC-MPC with the convexified tightened collision avoidance constraint to that with the direct implementation of the approximated constraint. To this end, first the same simulation setup as presented in Section \ref{subsec:direct} was used. As before, 100 simulations for each of the PA and US decoupling methods were completed. Here, when utilizing US decoupling, we combine US case 1 to find $\mu_{x_2^{'}}^*$ and US case 2 to find $\mu_{x_2^{'}}^*$, as noted in Section \ref{subsubsec:bb}. Figure \ref{fig:trajectoriesConv} shows the trajectories for the 100 simulations for both PA and US decoupling. When comparing this to Figure \ref{fig:trajectoriesProb}, we see that the convexified tightened constraint has significantly improved the robustness of our CC-MPC. Specifically, the CC-MPC was able to find a solution for all simulations. The ${ego}_i$ also passes each ${ov}_j$ on the same side in every simulation, which is required based on the formulation of the convexified tightened constraint.

\begin{figure}[!t]
\centering
\includegraphics[width=3.5in]{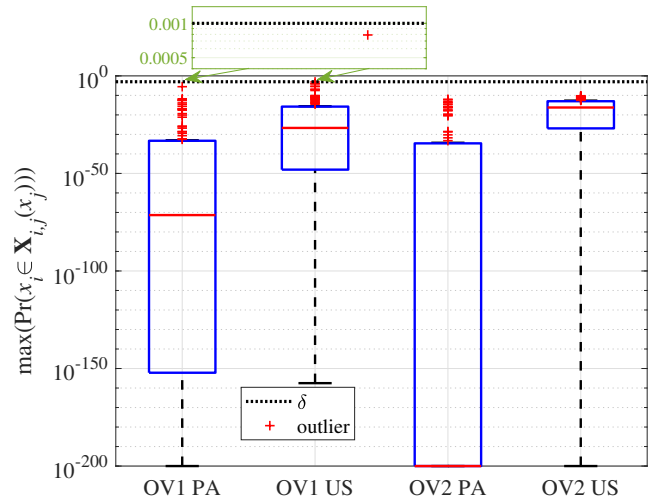}
\caption{Box plot of max observed probability of collision between the ${ego}_i$ and ${ov}_1$ and ${ov}_2$ for each simulation with the convexified tightened collision constraint.}
\label{fig:PrCboxPlotConv}
\end{figure}
The trajectories in Figure \ref{fig:trajectoriesConv} show improved robustness, however, we are also interested in whether the probability of collision threshold is not violated. The boxplot of the maximum observed probability of collision between each OV is presented in Figure \ref{fig:PrCboxPlotConv}. From the figure, we notice that the probability of collision threshold is never violated. The maximum observed probability of collision was approximately 0.0008, which occurred between the $ego_i$ and ${ov}_1$ with the US decoupling method. When comparing the two decoupling methods, it is easy to see in Figure \ref{fig:PrCboxPlotConv} that, as expected, the US decoupling method is significantly less conservative than the PA decoupling method. This is reflected in Figure \ref{fig:trajectoriesConv}, where the US trajectories deviate less from the centerline, traveling closer to the obstacles. Now comparing Figure \ref{fig:PrCboxPlotConv} with Figure \ref{fig:PrCboxPlotProb}, it is evident that the overall probability of collision distributions is shifted downward for the convexified constraint, i.e. the constraint convexification has resulted in safer driving on average.

\begin{figure}[!t]
\centering
\includegraphics[width=3.5in]{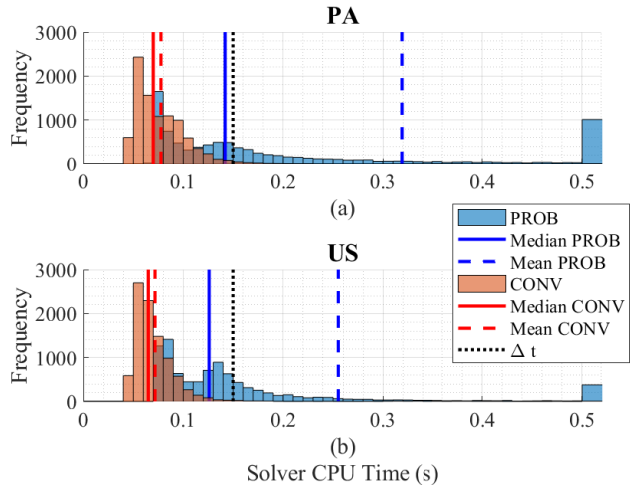}
\caption{Histogram of MPC computation time for (a) PA and (b) US decoupling; where $\Delta t$ is the simulation update rate, PROB denotes the direct implementation, and CONV denotes the convexified tightened constraint.}
\label{fig:cpuTimeHist}
\end{figure}
Next, it is important to determine if the reduction in conservatism comes at the cost of increased computation time. Figure \ref{fig:cpuTimeHist} presents the distributions of the computation times for both CC-MPC implementations and both decoupling methods. The distribution is significantly shifted to the left (or computation times are on average lower) for the convexified constraint. Specifically, using the convexified constraint, the median computation times are reduced to 0.07s for the PA decoupling case and 0.065s for the US decoupling case. With the convexified constraint, the real-time requirement ($\Delta_t$) is exceeded only in 1.85\% of the MPC calls for the PA case and in 1.31\% of the MPC calls for the US case.

The improved computation time does come at the cost of increased precomputations for finding the dimensions of the tightened constraint bounding box. The computation statistics for finding the bounding box dimensions are presented in Table \ref{table:BBcpuTime}. There is a negligible difference in the computation times for the two decoupling methods. Further, the magnitude of the additional precomputation time is significantly lower than the reduction in the CC-MPC execution time. We do note that the presented computation time is only for finding one set of tightened constraint bounding box dimensions. This process must be repeated for each step of the horizon, however, each computation is independent and can therefore be parallelized. Thus, the convexified tightened constraint is a viable solution to improving the robustness and reducing the execution time of the presented CC-MPC.
\begin{table}[!t]
\renewcommand{\arraystretch}{1.3}
\caption{Computation Time Statistics for Finding the Tightened Constraint Bounding Box Dimensions}
\label{table:BBcpuTime}
\centering
\begin{tabular}{|c||c|c|}
\hline
Computation Time (ms) & PA & US\\
\hline
Median & 3.999 & 4.035\\
\hline
Upper Quartile & 4.670 & 4.692\\
\hline
\end{tabular}
\end{table}

\begin{figure}[!t]
\centering
\includegraphics[width=4.5in]{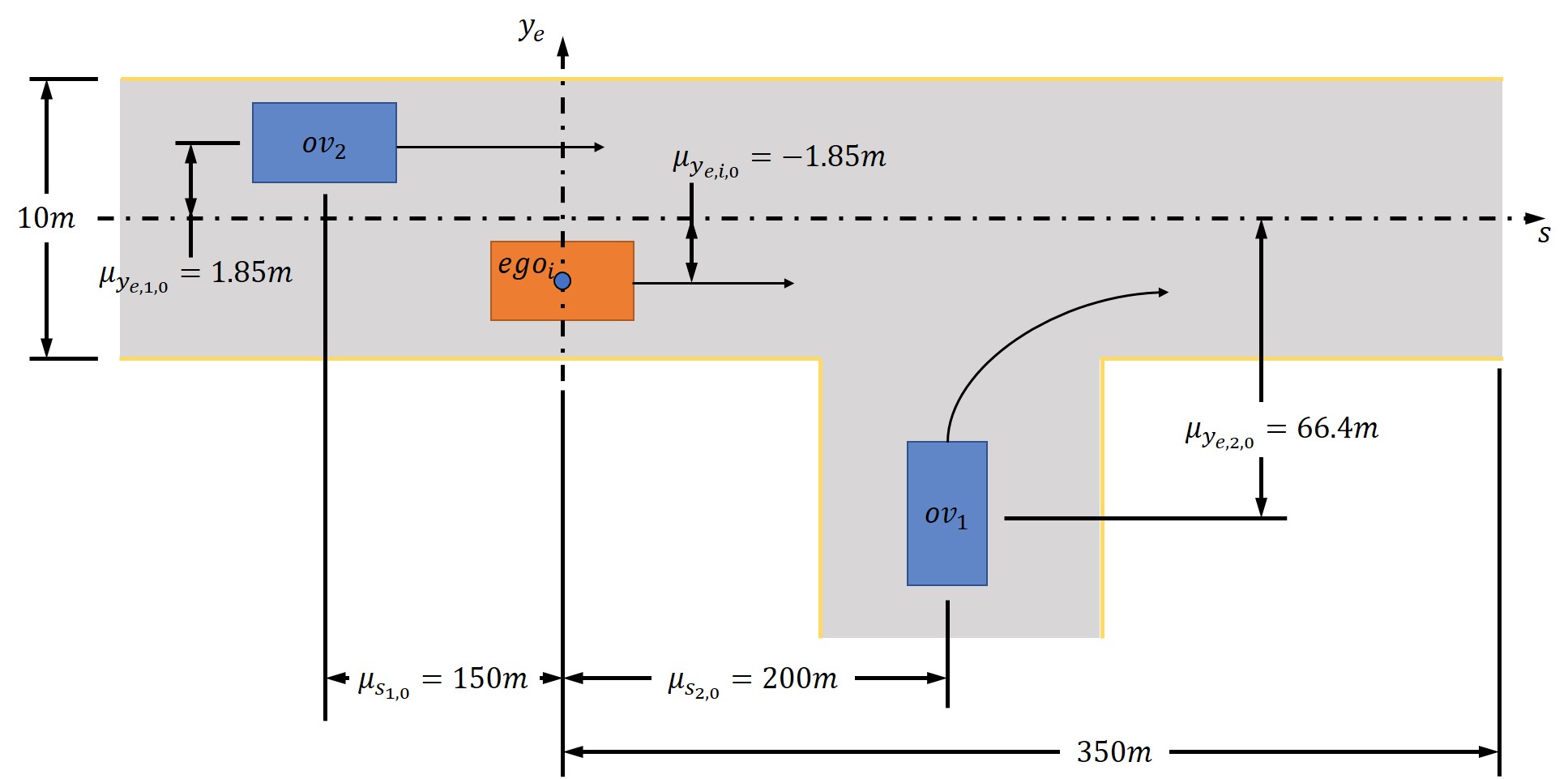}
\caption{Scenario to illustrate performance with moving obstacles. The OV references are $r_{v,1} = 8$ m/s and $r_{v,2} = 22$ m/s for speed and $r_{y_e,1} = -1.85$ m and $r_{y_e,2} = 1.85$ m for lateral error, respectively, while the ego references are $r_{v,i}=20$ m/s and $r_{y_e,i} = -1.85$ m.}
\label{fig:scenario2}
\end{figure}

Lastly, in order to show the applicability of the constraint handling method to more complex scenarios, we will briefly present results for a second scenario depicted in Figure \ref{fig:scenario2}.
Therein, the ego vehicle is traveling in the right lane, with $ov_2$ passing $ego_i$ on the left, while $ov_1$ is entering the roadway from the right in front of $ego_i$. The mean initial positions (of the randomize simulations) are shown in Figure \ref{fig:scenario2}, while the initial velocities are as follows: $v_{i,0} = 17$ m/s, $v_{1,0} = 15$ m/s, and $v_{2,0} = 20$ m/s. The OVs utilize the intelligent driver model (IDM) for speed tracking and a linear feedback controller for $y_e$ and $\psi$ (the reference heading error $r_\psi=0$ rad).

\begin{figure}[!t]
\centering
\includegraphics[width=3.5in]{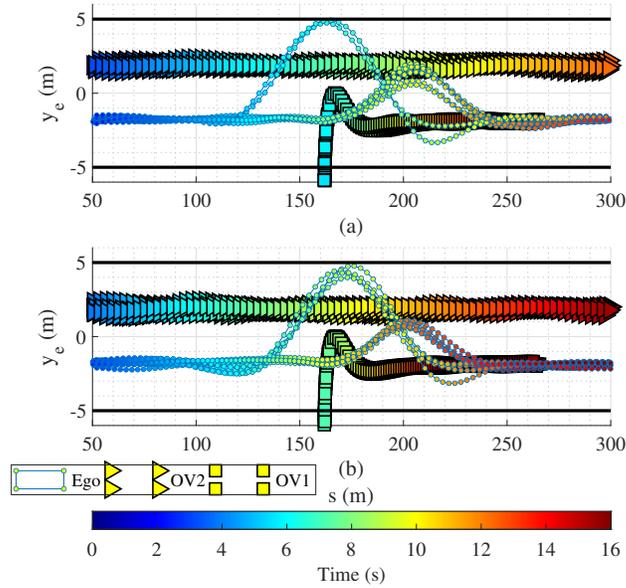}
\caption{Simulation trajectories for CC-MPC with convexified tightened collision constraints and scenario 2 using (a) PA and (b) US decoupling.}
\label{fig:trajConv2}
\end{figure}

\begin{figure}[!t]
\centering
\includegraphics[width=3.5in]{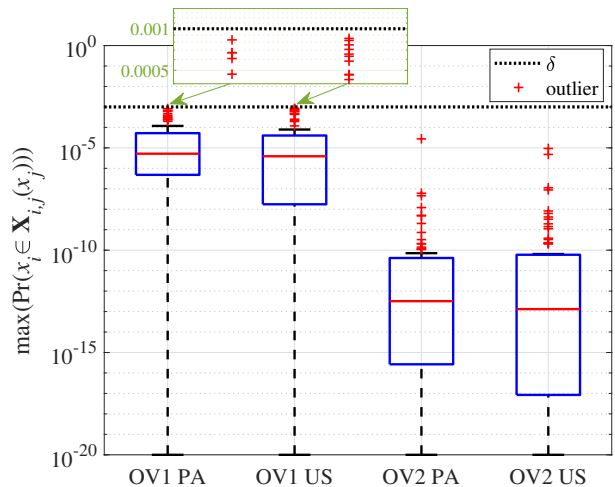}
\caption{Box plot of max observed probability of collision between the ${ego}_i$ and ${ov}_1$ and ${ov}_2$ for scenario 2 with the convexified tightened collision constraint.}
\label{fig:PrCboxPlotConv2}
\end{figure}

Figure \ref{fig:trajConv2} shows select simulated trajectories with the CC-MPC and the two decoupling methods. For clarity only representative trajectories are shown.
In general, there are two main types of responses that $ego_i$ exhibits, either: speeding up and passing $ov_1$ before returning to the left and allowing $ov_2$ to pass; or slowing down and allowing $ov_2$ to pass prior to passing $ov_1$. Despite the varied responses, the probability of collision criteria are still met as can be seen in Figure \ref{fig:PrCboxPlotConv2}.
When comparing to Figure \ref{fig:PrCboxPlotConv}, it is evident that with moving obstacles the maximum probability of collision distributions are shifted upwards as the dynamic environment includes additional uncertainties with respect to OV state predictions that are not captured with the simplified models used here. Still, the thresholds were not violated in any of the simulations.

\section{Conclusion}\label{sec:conclusion}
In this paper we have derived analytical methods for the efficient approximation of an upper bound on the mutual probability of collision between robotic vehicles. This included a detailed discussion of two separate decoupling methods (principal axes rotation, and unitary scaling cases 1 and 2) and the associated conservatism of each method. We have drawn the following important insights on the utility of either method: principal axis rotation is ideal for unstructured environments as the conservatism introduced is easily bounded in all directions; unitary scaling case 1 is ideal for environments where agents are traveling in parallel and passing each other, because the conservatism is only introduced in the longitudinal direction; lastly, unitary scaling case 2 is ideal for environments where agents are following each other as conservatism is only introduced in the lateral direction.

The effectiveness of the analytical approximation methods was then evaluated within a chance-constrained model predictive controller (CC-MPC) motion planner formulation. Therein, we highlighted shortcomings in the robustness and computational efficiency of using a direct implementation of the tightened probability of collision constraints in the CC-MPC. We then proposed a method for convexification of the tightened constraints which showed promising results for significantly improving robustness and reducing computation times for online motion planning in complex traffic.

There are important future extensions that can be pursued on the application of these approximation methods for mutual collision probabilities. These include the following: Verifying the performance when implemented on a higher fidelity plant/vehicle model; Using the approximations for clustering object vehicles in traffic to enable further convexification of the navigation space \cite{Wang2018}; Improving the algorithm presented here for determining the slope and intercept of the convexified tightened constraint; Exploring methods/extensions to the constraint building process to consider unstructured environments and roadways with multiple lanes; Investigating methods to account for the inherent coupling between neighboring vehicle's states with an additional layer of uncertainty decoupling.



\appendix
\section{Calculating $\Sigma_{x_{ij}^{'}}$ and $\theta$}\label{app:PAcalcs}

In order to implement principal axis rotation it is necessary to solve for the covariance $\Sigma_{x_{ij}^{'}}$ along the principal axes and the angle of rotation $\theta$ to align the reference frame with the principal axes. To accomplish this, we first solve for the eigenvalues of the original covariance $\Sigma_{x_{ij}}$, as $\Sigma_{x_{ij}^{'}}=D$, where $D$ is the diagonal matrix of eigenvalues. Assuming the covariance matrix is of the form:
\begin{equation}\label{eq:cov_def}
\Sigma_{x_{ij}}=\begin{bmatrix}\sigma_x^2&\rho\sigma_x\sigma_y\\
\rho\sigma_x\sigma_y&\sigma_y^2 \end{bmatrix},
\end{equation}
we can compute the eigenvalues $\lambda_c$, $c\in\left\{1,2\right\}$, by solving $\left|\Sigma_{x_{ij}}-\lambda_cI\right|=0$, resulting in:
\begin{equation}\label{eq:lambda}
\lambda_{1/2}=\frac{\left(\sigma_x^2+\sigma_y^2\right)\mp\sqrt{\left(\sigma_x^2+\sigma_y^2\right)^2-4\sigma_x^2\sigma_y^2\left(1-\rho^2\right)}}{2}.
\end{equation}
Then, the diagonal matrix $D=\begin{bmatrix}\lambda_1&0\\0&\lambda_2\end{bmatrix}$.

Now that we have the covariance along the principal axes, it is necessary to solve for the rotation angle $\theta$. We first solve for the matrix of eigenvectors $V=\begin{bmatrix}\vec{v}_1&\vec{v}_2\end{bmatrix}$, where $\vec{v}_c=\begin{bmatrix}v_{c1}&v_{c2}\end{bmatrix}^T$, for all $c\in\left\{1,2\right\}$, is the unit eigenvector corresponding to the $c^{th}$ eigenvalue. Knowing that $\left(\Sigma_{x_{ij}}-\lambda_c I\right)\vec{v}_c=0$ and $v_{c1}^2+v_{c2}^2=1$, as we assume unit eigenvectors, we can solve for $v_{c1}$  and $v_{c2}$ as follows:
\begin{subequations}\label{eq:vc1and2}
\begin{gather}
v_{c2}=\pm\left[\left(\frac{\sigma_x\sigma_y\rho}{\sigma_x^2-\lambda_c}\right)^2+1\right]^{-1/2};\label{eq:vc2}\\
v_{c1}=-\frac{\sigma_x\sigma_y\rho v_{c2}}{\sigma_x^2-\lambda_c}.\label{eq:vc1}
\end{gather}
\end{subequations}
Further, as the eigenvectors are orthogonal, it can be shown that $v_{21}=\mp v_{12}$ and $v_{22}=\pm v_{11}$. The angle of rotation $\theta_c$ is then:
\begin{equation}\label{eq:theta}
\theta_c=-\tan^{-1}{\frac{v_{c2}}{v_{c1}}}.
\end{equation}
Substituting \eqref{eq:vc1} in to \eqref{eq:theta} and simplifying yields:
\begin{equation}\label{eq:theta_simplified}
\theta_c=-\tan^{-1}{\frac{\lambda_c-\sigma_x^2}{\sigma_x\sigma_y\rho}}.
\end{equation}
For our implementation we rotate to align with the first principal axis or $\theta_1$. The rotation applied to $x_{ij}$ and $\mathbb{X}_{ij}\left(0,\Phi_{ji,l}\right)$ is then:
\begin{equation}\label{eq:TPA}
T_c=\begin{bmatrix}\cos{\theta_c} & -\sin{\theta_c}\\
\sin{\theta_c}&\cos{\theta_c}\end{bmatrix},
\end{equation}
where:
\begin{subequations}\label{eq:sinAndcos_expanded}
\begin{gather}
\cos{\theta_c}=\frac{\sigma_x\sigma_y\rho}{\sqrt{\left(\lambda_c-\sigma_x^2\right)^2+\left(\sigma_x\sigma_y\rho\right)^2}};\label{eq:costheta_expanded}\\
\sin{\theta_c}=\frac{\lambda_c-\sigma_x^2}{\sqrt{\left(\lambda_c-\sigma_x^2\right)^2+\left(\sigma_x\sigma_y\rho\right)^2}}.\label{eq:sintheta_expanded}
\end{gather}
\end{subequations}
Lastly, as we perform the inverse rotation in order to bring the tightened constraint bounding box back to the original reference frame for use within our motion planning algorithm, we note:
\begin{equation}\label{eq:TPAinv}
T_c^{-1}=\begin{bmatrix}\cos{-\theta_c}&-\sin{-\theta_c}\\
\sin{-\theta_c}&\cos{-\theta_c}\end{bmatrix}.
\end{equation}

\section{Unitray Scaling - Calculating $T_0$}\label{app:T0}

In order to efficiently compute the transformation matrix $T_0=\Sigma_{x_{ij}}^{-1/2}$, we can take advantage of the fact that the covariance matrix $\Sigma_{x_{ij}}$ is symmetric. Due to symmetry, it is possible to orthogonally diagonalize the covariance $\Sigma_{x_{ij}}=VDV^{-1}=VDV^T$ \cite{Lay2016}. See \ref{app:PAcalcs} for how to compute $D$ and $V$. The square root by diagonalization is then simply $\Sigma_{x_{ij}}^{1/2}=VD^{1/2}V^T$, where $D^{1/2}=\begin{bmatrix}\sqrt{\lambda_1}&0\\
0&\sqrt{\lambda_2}\end{bmatrix}$. To obtain $T_0=\left(VD^{1/2}V^T\right)^{-1}$, we take the inverse: 
\begin{equation}\label{eq:T0}
T_0=\\
\frac{1}{\left(v_{11}v_{22}-v_{21}v_{12}\right)^2\lambda_1^{\frac{1}{2}}\lambda_2^{\frac{1}{2}}}A,
\end{equation}
where:
\begin{multline}\label{eq:detSqrtSigma}
A=\\
\begin{bmatrix}v_{12}^2\lambda_1^{\frac{1}{2}}+v_{22}^2\lambda_2^{\frac{1}{2}}&-v_{11}v_{12}\lambda_1^{\frac{1}{2}}-v_{21}v_{22}\lambda_2^{\frac{1}{2}}\\
-v_{11}v_{12}\lambda_1^{\frac{1}{2}}-v_{21}v_{22}\lambda_2^{\frac{1}{2}}&v_{11}^2\lambda_1^{\frac{1}{2}}+v_{21}^2\lambda_2^{\frac{1}{2}}\end{bmatrix}.
\end{multline}

\section{Unitary Scaling - Decomposing $T_0$}\label{app:decompT0}
As mentioned previously, the transformation $T_0$ can be decomposed into three elementary operations: rotation, scaling, and shear. From a matrix perspective this means $T_0$ may be decomposed into $T_0=RSH$, where the rotation matrix $R$, scaling matrix $S$, and shear matrix $H$ are defined as:
\begin{subequations}
\begin{gather}
R=\begin{bmatrix}\cos{\alpha}&-\sin{\alpha}\\
\sin{\alpha}&\cos{\alpha}\end{bmatrix};\label{eq:R}\\
S = \begin{bmatrix}s_1&0\\
0&s_2\end{bmatrix};\label{eq:S}\\
H = \begin{cases}\bigg[\begin{matrix}1&h\\
0&1\end{matrix}\bigg]&\text{for case 1},\\
\bigg[\begin{matrix}1&0\\
h&1\end{matrix}\bigg]&\text{for case 2}.\end{cases}\label{eq:H}
\end{gather}
\end{subequations}
Multiplying $R$, $S$, and $H$, and setting the elements of $RSH$ equal to those of $T_0=\Big[\begin{matrix}t_{11}&t_{12}\\t_{12}&t_{22}\end{matrix}\Big]$ yields a system of 4 equations. We first detail the solution for case 1, followed by that for case 2.

\subsection{Case 1 - Longitudinal Conservatism}\label{subapp:case1}
The system of equations for case 1 is:
\begin{subequations}
\begin{gather}
t_{11}=s_1\cos{\alpha};\\
t_{12}=s_1h\cos{\alpha}-s_2\sin{\alpha};\\
t_{12}=s_1\sin{\alpha};\\
t_{22}=s_1h\sin{\alpha}+s_2\cos{\alpha},
\end{gather}
\end{subequations}
with 4 unknowns, $\alpha$, $s_1$, $s_2$, and $h$. Solving this yields:
\begin{subequations}\label{eq:RSHelcase1}
\begin{gather}
\alpha=\tan^{-1}{\frac{t_{12}}{t_{11}}};\\
s_1=\sqrt{t_{11}^2+t_{12}^2};\label{eq:s1case1}\\
s_2=\frac{t_{11}t_{22}-t_{12}^2}{\sqrt{t_{11}^2+t_{12}^2}};\label{eq:s2case1}\\
h=\frac{t_{12}\left(t_{22}+t_{11}\right)}{t_{12}^2+t_{11}^2}.\label{eq:hcase1}
\end{gather}
\end{subequations}
Recalling we seek the transformation matrix $T=SH$, it is beneficial to obtain \eqref{eq:s1case1}-\eqref{eq:hcase1} in terms of the eigenvalues and eigenvector elements. Substituting the definitions of $t_{11}$, $t_{12}$, and $t_{22}$ from \eqref{eq:T0} and \eqref{eq:detSqrtSigma} and simplifying yields:
\begin{subequations}\label{eq:s1s2hcase1}
\begin{gather}
s_1=\left(\frac{v_{12}^2\lambda_1+v_{11}^2\lambda_2}{\lambda_1\lambda_2}\right)^{\frac{1}{2}};\\
s_2=\left(v_{12}^2\lambda_1+v_{11}^2\lambda_2\right)^{\frac{1}{2}};\\
h=\frac{v_{11}v_{12}\left(\lambda_2-\lambda_1\right)}{v_{12}^2\lambda_1+v_{11}^2\lambda_2}.\label{eq:hOfvlambdaCase1}
\end{gather}
\end{subequations}

Now to see if this equation is unbounded we will look at the limits of \eqref{eq:hOfvlambdaCase1} with respect to the eigenvectors and eigenvalues. We know that the eigenvector elements are bounded by $\left[-1,1\right]$ and $v_{11}^2+v_{12}^2=1$, as we assumed unit eigenvectors. It is easy to see that if $v_{11}=\pm1$ and $v_{12}=0$, or vice versa, $h=0$. Therefore, we are interested in the limits with respect to the eigenvalues. Taking the limit of \eqref{eq:hOfvlambdaCase1} as $\lambda_2\to\pm\infty$ and using L’H\^{o}pital’s rule yields $v_{12}/v_{11}$, which is the tangent of the eigenvector and therefore unbounded. Similarly, taking the limit as $\lambda_1\to\pm\infty$ yields $-v_{11}/v_{12}$, which is the negative of the cotangent of the eigenvector and also unbounded. Although, the shear is theoretically unbounded, from a practical perspective, if you can characterize and bound the spectral radius of $\Sigma_{x_{ij}}$, the shear, and conservatism is therefore bounded. 

\subsection{Case 2 - Lateral Conservatism}\label{subapp:case2}

The system of equations for case 2 is:
\begin{subequations}
\begin{gather}
t_{11}=s_1\cos{\alpha}-s_2h\sin{\alpha};\\
t_{12}=-s_2\sin{\alpha};\label{eq:s1case2}\\
t_{12}=s_1\sin{\alpha}+s_2h\cos{\alpha};\label{eq:s2case2}\\
t_{22}=s_2\cos{\alpha}.\label{eq:hcase2}
\end{gather}
\end{subequations}
Solving the systems results in:
\begin{subequations}
\begin{gather}
\alpha=\tan^{-1}{-\frac{t_{12}}{t_{22}}};\\
s_1=\frac{t_{11}t_{22}-t_{12}^2}{\sqrt{t_{12}^2+t_{22}^2}};\\
s_2=\sqrt{t_{12}^2+t_{22}^2};\\
h=\frac{t_{12}\left(t_{22}+t_{11}\right)}{t_{12}^2+t_{22}^2}.
\end{gather}
\end{subequations}
As we did with case 1, we substitute in to \eqref{eq:s1case2}-\eqref{eq:hcase2} from \eqref{eq:T0} and \eqref{eq:detSqrtSigma} and simplify to obtain: 
\begin{subequations}\label{eq:s1s2hcase2}
\begin{gather}
s_1=\left(v_{11}^2\lambda_1+v_{12}^2\lambda_2\right)^{\frac{1}{2}};\\
s_2=\left(\frac{v_{11}^2\lambda_1+v_{12}^2\lambda_2}{\lambda_1\lambda_2}\right)^{\frac{1}{2}};\\
h=\frac{v_{11}v_{12}\left(\lambda_2-\lambda_1\right)}{v_{11}^2\lambda_1+v_{12}^2\lambda_2}.\label{eq:hOfvlambdaCase2}
\end{gather}
\end{subequations}

Comparing \eqref{eq:hOfvlambdaCase1} of case 1 and \eqref{eq:hOfvlambdaCase2} of case 2, we can see the only difference is the eigenvalues are weighted by the opposite elements of the eigenvector in the denominator. The results for analyzing the limits with respect to $v_{11}$ and $v_{12}$, therefore, still hold. However, the limit of $h$ as $\lambda_1\to\pm\infty$ is now $-v_{12}/v_{11}$, and the limit of $h$ as $\lambda_2\to\pm\infty$ is $v_{11}/v_{12}$. Again, this is unbounded.

\section{Probability Threshold Search Algorithm}\label{app:PTS}

The probability threshold search (PTS) algorithm is used within step \ref{bb_step3} of the process in Section \ref{subsubsec:bb} for finding the tightened constraint minimum bounding box. PTS (Algorithm \ref{alg:PTS}) is based on Newton’s method.

\begin{algorithm}[H]
\caption{Probability Threshold Search (PTS) Algorithm}
\label{alg:PTS}
\begin{algorithmic}[1]
\Require Threshold $\delta$, maximum iterations $\overline{N}_i$, tolerance $\varepsilon$, initial guess $\mu_{x_{2,0}'}$, $\left\{\Pr\left(\phi_{ji}\in\Phi_{ji,l}\right),a_l^{''},b_l^{''}\right\}\forall l \in\left\{1,...,n_\phi\right\}$, $\Sigma_{x_{ij}'}$, and $\Pr\left(x_1^{'}\in\left[\underline{x}_{1,l}^{'},\overline{x}_{1,l}^{'}\right]\right)$ 
\Ensure Distance $\mu_{x_2'}^*$, Probability $\Pr\left(x_{ij}^{'}\in\mathbb{X}_{ij}^{''}\left(0\right)\right)$, Iteration number $i$
\State \textbf{Initialize} $\mu_{x_2'}^*=\mu_{x_{2,0}'}$, $i=1$, $\underline{\mu}_{x_2'}=0$, $\overline{\mu}_{x_2'}=1e5$
\While{$i\leq\overline{N}_l$}
\State Calculate $\Pr\left(x_{ij}^{'}\sim\mathcal{N}\left(\begin{bmatrix}0\\\mu_{x_2'}^*\end{bmatrix},\Sigma_{x_{ij}'}\right)\in\mathbb{X}_{ij}^{''}\left(0\right)\right)$
\State Calculate deviation to $\delta$: $dP=\Pr\left(x_{ij}^{'}\in\mathbb{X}_{ij}^{''}\left(0\right)\right)-\delta$
\If{$\left|dP\right|\leq\varepsilon$}
\State \textbf{break while}
\EndIf
\If{$dP<0$ and $\mu_{x_2'}^*<\overline{\mu}_{x_2'}$}
\State $\overline{\mu}_{x_2'}=\mu_{x_2'}^*$
\EndIf
\If{$dP>0$ and $\mu_{x_2'}^*>\underline{\mu}_{x_2'}$}
\State $\underline{\mu}_{x_2'}=\mu_{x_2'}^*$
\EndIf
\State Calculate the derivative $\frac{\partial}{\partial\mu_{x_2'}^*}\Pr\left(x_{ij}^{'}\in\mathbb{X}_{ij}^{''}\left(0\right)\right)$
\If{$\mu_{x_2'}^*-dP/\frac{\partial}{\partial\mu_{x_2'}^*}\Pr\left(x_{ij}^{'}\in\mathbb{X}_{ij}^{''}\left(0\right)\right)>\overline{\mu}_{x_2'}$}
\State $\mu_{x_2'}^*=\bigl(\overline{\mu}_{x_2'}+\underline{\mu}_{x_2'}\bigr)/2$
\ElsIf{$\mu_{x_2'}^*-dP/\frac{\partial}{\partial\mu_{x_2'}^*}\Pr\left(x_{ij}^{'}\in\mathbb{X}_{ij}^{''}\left(0\right)\right)<\underline{\mu}_{x_2'}$}
\State $\mu_{x_2'}^*=\bigl(\overline{\mu}_{x_2'}+\underline{\mu}_{x_2'}\bigr)/2$
\Else 
\State $\mu_{x_2'}^*=\mu_{x_2'}^*-dP/\frac{\partial}{\partial\mu_{x_2'}^*}\Pr\left(x_{ij}^{'}\in\mathbb{X}_{ij}^{''}\left(0\right)\right)$
\EndIf
\State $i=i+1$
\EndWhile
\end{algorithmic}
\end{algorithm}

\bibliographystyle{elsarticle-num} 
\bibliography{elsarticle-CT_Journal}


%
%
%
\end{document}